\documentclass[journal]{IEEEtran}
\usepackage{cite}
\usepackage{graphicx}
\usepackage{amsfonts}
\usepackage{amsmath}
\usepackage{amssymb}
\usepackage{diagbox}
\usepackage{multirow}
\usepackage{graphics}
\usepackage{epstopdf}
\usepackage{tabularx}
\usepackage{epsfig}
\usepackage{color,subfigure}
\usepackage{graphicx}
\usepackage{setspace}
\usepackage{array}
\usepackage[utf8]{inputenc} 
\usepackage[T1]{fontenc}    
\usepackage{hyperref}       
\usepackage{url}            
\usepackage{booktabs}       
\usepackage{amsfonts}       
\usepackage{nicefrac}       
\usepackage{microtype}      
\usepackage{xcolor}         
\usepackage{times}
\usepackage{epsfig}
\usepackage{graphicx}
\usepackage{amsmath}
\usepackage{amssymb}
\usepackage{wrapfig}
\usepackage{amsmath}
\usepackage{amssymb}
\usepackage{algorithm}
\usepackage{algorithmic}
\usepackage{multirow}
\usepackage{url}

\ifCLASSINFOpdf
\else
\fi
\begin{document}
%
\title{Boosting Fast
Adversarial Training with Learnable Adversarial
Initialization}
%
%
%

%
\author{Xiaojun Jia,
        Yong Zhang,
        Baoyuan Wu,
        Jue Wang
        and~Xiaochun Cao

\IEEEcompsocitemizethanks{
\IEEEcompsocthanksitem Manuscript received October 11, 2021; returned March 23, 2022;
revised May 4, 2022; accepted May 29, 2022. This work was supported
by the National Key R\&D Program of China under Grant 2019YFB1406500, National Natural Science Foundation of China (No. 62025604, 62132006, U1803264, 62076213), Open Project Program of State Key Laboratory of Virtual Reality Technology and Systems, Beihang University (No.VRLAB2021C06), Tencent AI Lab Rhino-Bird Focused Research Program under grant No. JR202123. The associate editor
coordinating the review of this manuscript and approving it for publication
was Prof. Jiaying Liu. (Corresponding author: Yong Zhang and Xiaochun Cao.)
\IEEEcompsocthanksitem Xiaojun Jia is with
State Key Laboratory of Information Security,
Institute of Information Engineering, Chinese Academy of Sciences, Beijing 100093, China, and also with School of Cyber Security, University of Chinese Academy of Sciences, Beijing 100049, China.
(e-mail: jiaxiaojun@iie.ac.cn)
\IEEEcompsocthanksitem Yong Zhang and Jue Wang are with AI Lab, Tencent Inc., Shenzhen 518000, China.(e-mail:
  $\{$zhangyong201303, arphid$\}$@gmail.com)
\IEEEcompsocthanksitem Baoyuan Wu is with School of Data Science, the Chinese University of Hong Kong, Shenzhen (CUHK-Shenzhen) and Secure Computing Lab of Big Data, Shenzhen Research Institute of Big Data (SBRID), Shenzhen 518172, China. 
(e-mail: wubaoyuan@cuhk.edu.cn)
\IEEEcompsocthanksitem  Xiaochun Cao is with School of Cyber Science and Technology, Shenzhen Campus, Sun Yat-sen University, Shenzhen 518107, China (e-mail: caoxiaochun@mail.sysu.edu.cn)
}
}

%
%

\markboth{Manuscript for IEEE Transactions on Image Processing}%
{Shell \MakeLowercase{\textit{et al.}}: Bare Demo of IEEEtran.cls for IEEE Journals}
%



\maketitle


\begin{abstract}
Adversarial training (AT) has been demonstrated to be effective in improving model robustness by leveraging adversarial examples for training.
However, most AT methods are in face of expensive time and computational cost for calculating gradients at multiple steps in generating adversarial examples. 
To boost training efficiency, fast gradient sign method (FGSM) is adopted in fast AT methods by calculating gradient only once.
Unfortunately, the robustness is far from satisfactory. 
One reason may arise from the initialization fashion. 
Existing fast AT generally uses a random sample-agnostic initialization, which facilitates the efficiency yet hinders a further robustness improvement.
Up to now, the initialization in fast AT is still not extensively explored.
In this paper, {focusing on image classification,} we boost fast AT with a sample-dependent adversarial initialization, \textit{i.e.}, an output from a generative network conditioned on a benign image and its gradient information from the target network.  
As the generative network and the target network are optimized jointly in the training phase, the former can adaptively generate an effective initialization with respect to the latter, which motivates gradually improved robustness. 
Experimental evaluations on four benchmark databases demonstrate the superiority of our proposed method over state-of-the-art fast AT methods, as well as comparable robustness to advanced multi-step AT methods. The code is released at \textcolor{red}{\textit{\url{https://github.com//jiaxiaojunQAQ//FGSM-SDI}}}.
\end{abstract}

\section{Introduction}
\IEEEPARstart{D}{eep} neural networks (DNNs) have successfully made breakthroughs in many fields, especially in image recognition \cite{krizhevsky2012imagenet, sainath2013deep}, and speech recognition \cite{deng2013new, amodei2016deep}. However, DNNs have been found vulnerable to adversarial examples \cite{szegedy2013intriguing}, \textit{i.e.,} the examples added with imperceptible perturbations can easily fool well-trained DNNs.
It has been proven that DNNs' real-world applications \cite{goodfellow2014explaining, zou2019reinforced,wei2018transferable,baiimproving,DBLP:journals/tip/CheBZLLTGC21} are also vulnerable to adversarial examples. Adversarial examples thus pose a huge threat to the commercial applications of deep learning. Improving the model robustness against adversarial examples is a challenging and important issue. A series of defense methods \cite{song2017pixeldefend,madry2017towards, liao2018defense,jia2019comdefend, dai2021deep,zhang2021defense,Bai21, DBLP:conf/nips/WuX020, DBLP:journals/tip/ZouSSY21,DBLP:journals/tip/ZhaoWSYHH21,DBLP:journals/tip/LiuLYZLT21 } have been proposed since then, among which adversarial training (AT) \cite{madry2017towards} has been proved to be among the most effective ones by injecting adversarial examples during training.  {In detail, the AT methods adopt the model gradient information to generate adversarial perturbation and then add the generated adversarial perturbation to the original clean sample to generate adversarial examples.} AT methods can be formulated as a minimax problem \cite{madry2017towards,shaham2018understanding} with the inner maximization maximizing the loss to generate adversarial examples and the outer minimization minimizing the loss on the generated adversarial examples to obtain a robust model. 
The robustness depends on the inner maximization \cite{wang2019convergence}, \textit{i.e.,} the adversarial example generation. But the generation of the adversarial examples is an NP-hard problem \cite{katz2017reluplex, weng2018towards}. Thus AT methods always adopt the model's gradient information to generate adversarial examples \cite{madry2017towards, szegedy2013intriguing}.
\par Based on the number of steps in generating adversarial examples, AT can be roughly categorized into two groups, \textit{i.e.,} multi-step AT \cite{madry2017towards,zhang2019theoretically,rice2020overfitting,lee2020adversarial}, and fast AT~\cite{tramer2017ensemble, shafahi2019adversarial, wong2020fast, andriushchenko2020understanding, kim2020understanding}.
Multi-step AT adopts multi-step adversarial attack methods such as projected gradient descent (PGD) \cite{madry2017towards} and achieves comprehensive robustness in defending against various attack methods.
However, they require a high computational cost to perform multiple forward and backward propagation calculations in generating adversarial examples. To boost training efficiency, fast AT methods are proposed, which need to calculate gradient only once and adopt fast gradient sign method (FGSM) \cite{szegedy2013intriguing}. 
Although they can greatly reduce time and computational cost, the robustness is far from satisfactory, compared with other state-of-the-art multi-step AT methods. 
Therefore, plenty of studies have explored how to improve the robustness of fast AT.
Among them, some studies \cite{wong2020fast, andriushchenko2020understanding} focus on the initialization issue, as it is proved that using a random initialization in fast AT plays an important role in improving robustness ~\cite{andriushchenko2020understanding}. 
However, the diverse random initialization fashions adopted in existing fast AT methods are usually sample-agnostic, which restricts further robustness improvement.
\begin{figure}
\begin{center}
\includegraphics[width=1.0\linewidth]{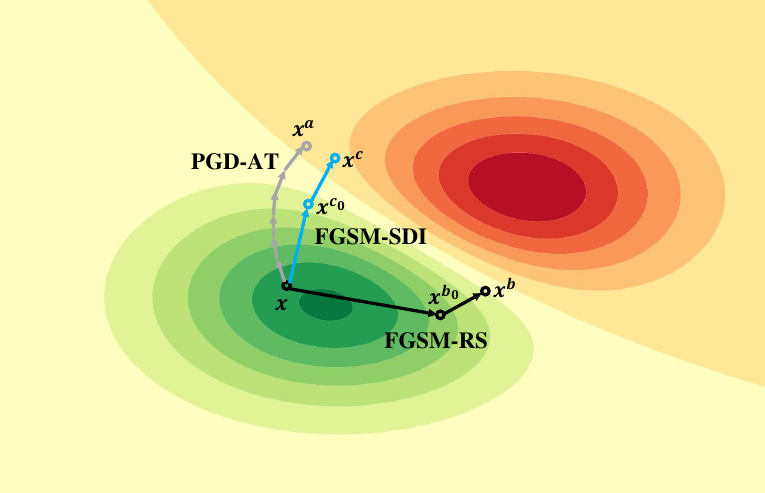}
\end{center}
\caption{Adversarial example generation process of PGD-AT~\cite{madry2017towards}, FGSM-RS~\cite{shafahi2019adversarial}, and our FGSM-SDI in the loss landscape of binary classification. Background is the contour of cross entropy. The redder the color, the lower the loss. PGD-AT is a multi-step AT method that computes gradients w.r.t the input at each step. FGSM-RS uses a random sample-agnostic initialization followed by FGSM, requiring the computation of gradient only once. But our FGSM-SDI uses a sample-dependent learnable initialization followed by FGSM. 
}
\label{fig:teaser}
\end{figure}

\par To overcome such a shortcoming, in this paper, {focusing on image classification,} we propose a sample-dependent  adversarial initialization to boost FGSM-based fast AT, dubbed \textit{FGSM-SDI}.
The sample-dependent initialization is calculated by a generative network conditioned on not only a benign image {which refers to the original clean image without adversarial perturbations}, but also its signed gradient from the target network. 
The benign image provides position information in the loss landscape, while the signed gradient provides a rough direction of increasing the loss. 
The initialization is then exploited by FGSM to generate a final adversarial example for training. 
The pipeline of adversarial example generation of our FGSM-SDI is illustrated in Fig.~\ref{fig:pipeline}. 
Note that the generative network and the target network are jointly learned under a minimax optimization framework, where the generative network attempts to create an effective initialization for FGSM to fool the target network while the target network improves its robustness against such adversarial examples via adversarial training.
Along with an increasingly robust target network, the generative network dynamically optimizes a more effective initialization for FGSM, boosting the robustness of fast AT.

Fig.~\ref{fig:teaser} presents the differences between a typical AT method (PGD-AT~\cite{madry2017towards}), a fast AT method (FGSM-RS~\cite{shafahi2019adversarial}), and our FGSM-SDI in generating adversarial examples. 
PGD-AT is an advanced multi-step AT method that can achieve decent performance in robustness but is time-consuming in calculating gradient at multiple steps.  
FGSM-RS calculates gradient only once at $x^{b_0}$ to which the benign image $x$ is moved with a random sample-agnostic initial perturbation.   
Differently, our FGSM-SDI calculates gradient at $x^{c_0}$ to which the benign image $x$ is moved with an optimized sample-dependent initial perturbation. 
Our learnable initialization depends on the benign image as well as its signed gradient, which is more informative than the random one. {Specifically, we adopt a lightweight generative network to generate the different adversarial initialization for different sample inputs, \textit{i.e.,} sample-specific initialization. And the generative network uses not only sample information but also sample gradient information to generate the adversarial initialization. In this way, compared with the random initialization, the proposed initialization is more informative}.
For PGD-AT, our FGSM-SDI can achieve comparable robustness with a much more efficient training process.
Compared to current fast AT methods (\textit{e.g.}, FGSM-RS), our FGSM-SDI outperforms them in robustness by a large margin, though with a slight sacrifice on efficiency due to the additional generative network. Note that such an additional generative network is in fact lightweight and acceptable. (see results in Sec.~\ref{sec:sota})
Our main contributions are summarized as follows:
\begin{itemize}
 \item We propose a sample-dependent adversarial initialization method for fast AT. The sample-dependent property is achieved by a generative network trained with both benign examples and their gradient information from the target network, which outperforms other sample-agnostic fast AT methods.
Our proposed adversarial initialization is dynamic and  optimized by the generative network along with the adjusted robustness of the target network in the training phase, which further enhances adversarial robustness.
 \item Extensive experiment results demonstrate that our proposed method not only shows a satisfactory training efficiency but also greatly boosts the robustness of fast AT methods. That is, it can achieve superiority over state-of-the-art fast AT methods, as well as comparable robustness to advanced multi-step AT methods.

\end{itemize}

\section{Related Work}
In this section, we first introduce the related researches on attack methods. Then we introduce the related researches on defense methods, especially the adversarial training variants. 
Specifically, in this paper, we focus on the image classification task, 
where adversarial examples can fool a well-trained image classification model into outputting the erroneous prediction with a high level of confidence. Given a clean image $x$ with the corresponding true label $y$ and a well-trained image classifier $f(\cdot)$, the attack methods are used to generate the adversarial example $x_{adv}$, to deceive the classifier into outputting an erroneous prediction, \textit{i.e.,} $f(x_{adv}) \neq f(x)=y$, where the distance function satisfies $L_{p}(x_{adv},x) \leq \epsilon$, where $\epsilon$ represents the maximum perturbation strength and $L_{p}$ represents the distance between the adversarial image $x_{adv}$ and the clean image $x$ under the $L_{p}$  distance metric, where $p \in\{1,2, \infty\}$.
In the recent researches of attack methods, $L_{ \infty}$ is a commonly used distance metric, which is also adopted in our paper. 

\subsection{Attack Methods}
Szegedy \textit{et al.}~\cite{szegedy2013intriguing} discover the existence of adversarial examples for DNNs and adopt a box-constrained L-BFGS method to generate adversarial examples. Goodfellow \textit{et al.}~\cite{goodfellow2014explaining} propose fast gradient sign method (FGSM) to generate adversarial examples. FGSM calculates the gradient of the loss function only once and then adds it to clean images to generate adversarial examples. Next, Madry \textit{et al.}~\cite{madry2017towards} propose Projected Gradient Descent (PGD) to generate adversarial examples. PGD iterates multiple times to perform a gradient descent step in the loss function to generate adversarial examples. Moosavi-Dezfooli \textit{et al.}~\cite{moosavi2016deepfool} propose DeepFool to efficiently generate adversarial examples for fooling deep networks. Then, DeepFool generates minimal adversarial perturbations that can fool the deep neural networks based on an iterative linearization of the classifier. And  Carlini \textit{et al.}~\cite{carlini2017towards}, propose a stronger attack method C\&W. C\&W introduces auxiliary variables to generate adversarial perturbations. After that, a series of iterative attack methods \cite{dong2018boosting, lin2019nesterov,xie2019improving,dong2019evading, wang2021enhancing} based on FGSM have been proposed to generate transferable adversarial examples. These attack methods focus on improving the adversarial transferability of adversarial examples, \textit{i.e.,} adversarial examples generated from one model can still be adversarial to another model. Recently, Croce F \textit{et al.}~\cite{croce2020reliable} propose two parameter-free attack methods, \textit{i.e.,} auto PGD with cross-entropy (APGD-CE) and auto PGD with the difference of logits ratio (APGD-DLR), to overcome the problem caused by the suboptimal step size and the objective function. Moreover, they combine the proposed attack methods with two existing attack methods, \textit{i.e.,} FAB \cite{croce2020minimally} and Square Attack \cite{andriushchenko2020square} to form the ensemble AutoAttack (AA). Furthermore, AA has achieved state-of-the-art performance in evaluating the model robustness against adversarial examples.


\subsection{Adversarial Training Methods}
Adversarial training (AT) variants have been widely accepted to improve adversarial robustness under comprehensive evaluations.
They can be formulated as a minimax optimization problem, \textit{i.e.,} the inner maximization maximizes the classification loss to generate adversarial examples and the outer minimization minimizes the loss of generated adversarial examples to train parameters of a robust model. 
Given a target network $f(\cdot, \mathbf{w})$ with parameters $\mathbf{w}$, a data distribution $\mathcal{D}$ including the benign sample $x$ and its corresponding label $y$, a loss function $\mathcal{L}(f(x, \mathbf{w}), y)$, and a threat bound $\triangle$, the objective function of AT can be defined as:
\begin{equation} 
\min_{\mathbf{w}} \mathbb{E}_{(x, y) \sim \mathcal{D}}\left[\max _{\delta \in \triangle} \mathcal{L}(f(x + \delta; \mathbf{w}), y)\right], \label{Eq:sat}
\end{equation}
where the threat bound can be defined as 
$\triangle=\{ \delta: \|\delta\| \leq \epsilon \}$ with the maximal perturbation intensity $\epsilon$.
The core of the adversarial training is how to find a better adversarial perturbation $\delta$. 
Typical adversarial training methods usually adopt a multi-step adversarial attack to generate an adversarial perturbation $\delta$, \textit{i.e.,} multiple steps of projected gradient ascent (PGD) \cite{madry2017towards}. It can be defined as: 
\begin{equation} \label{eq:pgd}
\delta_{t+1} =\Pi_{[-\epsilon, \epsilon]^{d}} [ \delta_{t}+ \alpha \text{sign}(\nabla_{x}\mathcal{L}(f(x + \delta_{t}; \mathbf{w}), y) ) ],
\end{equation}
where $\Pi_{[-\epsilon, \epsilon]^{d}}$ represents the projection to $[-\epsilon, \epsilon]^{d}$ and $\delta_{t+1}$ represents the adversarial perturbation after $t+1$ iteration steps.
In general, more iterations can boost robustness in adversarial training due to generating stronger adversarial examples. The prime PGD-based adversarial training framework is proposed in \cite{madry2017towards}. Following this framework, a larger number of PGD-based AT methods are proposed, amongst which an early stopping version \cite{rice2020overfitting} stands out. 
The algorithm of PGD-AT variants is summarized in Algorithm \ref{alg:PGD_AT}.

\noindent
\textbf{Fast Adversarial Training.}
Fast adversarial training variants are proposed recently by adopting the one-step fast gradient sign method (FGSM) \cite{goodfellow2014explaining}, which are also dubbed FGSM-based AT.
It can be defined as:

\begin{equation}
\delta^{*} =\epsilon \text{sign}(\nabla_{x}\mathcal{L}(f(x; \mathbf{w}), y) ) ,
\end{equation}
where $\epsilon$ represents the maximal perturbation strength. 
Although it accelerates adversarial training, it severely damages the robustness of the model, \textit{i.e.,} it cannot defend against the PGD attack {~\cite{andriushchenko2020understanding,kim2020understanding}}. 
FGSM-based AT has a catastrophic overfitting issue, \textit{i.e.,} the target model suddenly loses the robustness accuracy of adversarial examples generated by PGD (on the training data) during training.
Fortunately, Wong \emph{et al.}~ \cite{wong2020fast} propose to use FGSM-based AT combined with a random initialization to relieve the catastrophic overfitting. It can achieve  comparable performance to the prime PGD-based AT \cite{madry2017towards}. Specifically, they perform FGSM with a random initialization $\eta \in \mathbf{U}(-\epsilon ,\epsilon)$, where $\mathbf{U}$ represents a uniform distribution, which can be called FGSM-RS.
It can be defined as:
\begin{equation} \label{eq:rs_perturbation}
\delta^{*} =\Pi_{[-\epsilon, \epsilon]^{d}} [ \eta+ \alpha \text{sign}(\nabla_{x}\mathcal{L}(f(x +\eta; \mathbf{w}), y) ) ],
\end{equation}
where $\alpha$ represents the step size, which is set to $1.25\epsilon$ in \cite{wong2020fast}. This work demonstrates that combined with a good initialization, FGSM-based AT can achieve excellent performance as PGD-AT \cite{madry2017towards}. More importantly, compared with PGD-AT, the FGSM-RS requires a lower computational cost. The FGSM-RS algorithm is summarized in Algorithm \ref{alg:FGSM_AT}. 
Moreover, following FGSM-RS, several works are proposed to improve model robustness. Andriushchenko \emph{et al.}~\cite{andriushchenko2020understanding} find using a random initialization does not completely solve the catastrophic overfitting issue and propose a regularization method, dubbed FGSM-GA, to improve the performance of FGSM-based AT. Moreover,  Kim \emph{et al}. \cite{kim2020understanding} propose a stable single-step adversarial training based on FGSM-RS, \textit{a.k.a.,} FGSM-CKPT. FGSM-CKPT determines an appropriate magnitude of the perturbation for each image and thus prevents catastrophic overfitting.
 
\begin{algorithm}[t]
\caption{PGD-AT}
\label{alg:PGD_AT}
\begin{algorithmic}[1] 
\REQUIRE
The epoch $N$, the maximal perturbation $\epsilon$, the step size $\alpha$, the attack iteration $T$, the dataset $\mathcal{D}$ including the benign sample $x$ and the  corresponding label $y$, the dataset size $M$ and the network $f(\cdot, \mathbf{w})$ with parameters $\mathbf{w}$.

\FOR{$n=1,...,N$}
\FOR{$i=1,...,M$}
\STATE $\delta_{1} =0 $
\FOR{$t=1,...,T$}
\STATE $\delta_{t+1} =\Pi_{[-\epsilon, \epsilon]^{d}} [ \delta_{t}+ \alpha \text{sign}(\nabla_{x_{i}}\mathcal{L}(f(x_{i} + \delta_{t}; \mathbf{w}), y_{i}) ) ]$
\ENDFOR
\STATE $ \mathbf{w} \leftarrow \mathbf{w} -\nabla_{\mathbf{w}}\mathcal{L}(f(x_{i} + \delta_{t}; \mathbf{w}), y_{i})  $
\ENDFOR
\ENDFOR
\end{algorithmic}
\end{algorithm}
\begin{algorithm}[t]
\caption{FGSM-RS}
\label{alg:FGSM_AT}
\begin{algorithmic}[1] 
\REQUIRE
The epoch $N$, the maximal perturbation $\epsilon$, the step size $\alpha$, the dataset $\mathcal{D}$ including the benign sample $x$ and the  corresponding label $y$, the dataset size $M$ and the network $f(\cdot, \mathbf{w})$ with parameters $\mathbf{w}$.

\FOR{$n=1,...,N$}
\FOR{$i=1,...,M$}
\STATE $\eta =\mathbf{U}(-\epsilon ,\epsilon) $
\STATE $ \delta =\Pi_{[-\epsilon, \epsilon]^{d}} [ \eta+ \alpha \text{sign}(\nabla_{x}\mathcal{L}(f(x_{i} +\eta; \mathbf{w}), y) ) ]$
\STATE $ \mathbf{w} \leftarrow \mathbf{w} -\nabla_{\mathbf{w}}\mathcal{L}(f(x_{i} + \delta; \mathbf{w}), y_{i})  $
\ENDFOR
\ENDFOR
\end{algorithmic}
\end{algorithm}
 

\section{The Proposed Method}
\begin{figure*}
\begin{center}
 \includegraphics[width=1\linewidth]{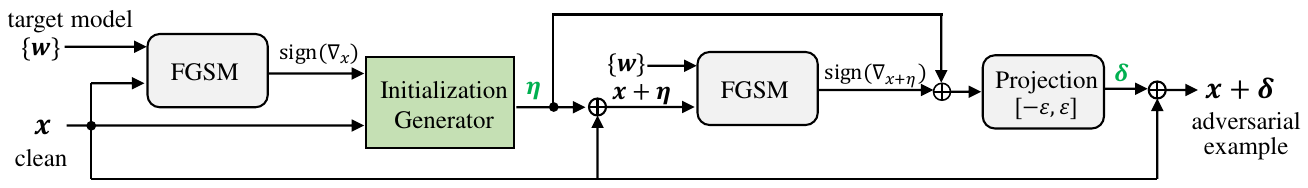}
\end{center}
\vspace{-3mm}
\caption{Adversarial example generation of the proposed FGSM-SDI. {The first FGSM is conducted  on the clean image for the initialization generator to generate the initialization. The second FGSM is performed on the input image  added with the generated initialization to
generate adversarial examples. The two FGSM modules keep the same in the FGSM-SDI. } }
\label{fig:pipeline}
\vspace{-3mm}
\end{figure*}
For fast AT, using a random sample-agnostic initialization is common and facilitates the efficiency, yet it hinders a further model robustness improvement. To remedy this issue, we propose a sample-dependent adversarial initialization to improve the robustness of fast AT as well as to overcome the catastrophic overfitting issue. The pipeline of the proposed method is introduced in Sec.~\ref{sec:pipeline},
the architecture of the proposed generative network is introduced in Sec.~\ref{sec:architecture} 
and the formulation is introduced in Sec.~\ref{sec:formulation}. 

\subsection{Pipeline of the Proposed Method} 
\label{sec:pipeline}
The proposed method consists of two networks, \textit{i.e.,} a generative network and a target network. The former one learns to produce a dynamic sample-dependent adversarial initialization for FGSM to generate adversarial examples, instead of using a random initialization. 
And the latter adopts the generated adversarial examples for training to improve model robustness.
As shown in Fig.~\ref{fig:pipeline}, a benign image and its gradient information from the target network are fed to the generative network and the generative network generates a sample-dependent initialization. 
FGSM is then performed on the input image added with the generated initialization to generate adversarial examples.
The target network is trained on the adversarial examples to improve the robustness against adversarial attacks.

For the target network, we adopt the architecture of a typical image classification network, defined as $y =f(x;\mathbf{w})$, where $x$ represents an input image, $y$ represents the predicted label, and $\mathbf{w}$ represents the parameters of the target network.

The generative network consists of three layers. The detailed structure of the generative network is presented in Sec.~\ref{sec:architecture}.
The inputs of the generative network are the benign image and its signed gradient. The signed gradient can be calculated as:
\begin{equation}
 s_x=\text{sign}(\nabla_{x}\mathcal{L}(f(x; \mathbf{w}), y) ),
\end{equation}
where $x$ is the input image and $y$ is the ground-truth label. The initialization generation process can be defined as:
\begin{equation} \label{eq:inti}
 \eta_{g} = \epsilon g(x,s_x;\theta),
\end{equation}
where $g(\cdot;\theta)$ represents the generative network with the parameters $\theta$, and $\eta_{g}$ represents the generated adversarial initialization. 
The output pixel value space of $g(\cdot;\theta)$ is $[-1,1]$. $\epsilon$ is a scale factor that maps the value to the range of $[-\epsilon, \epsilon]$. 
\begin{figure*}
\begin{center}
 \includegraphics[width=0.7\linewidth]{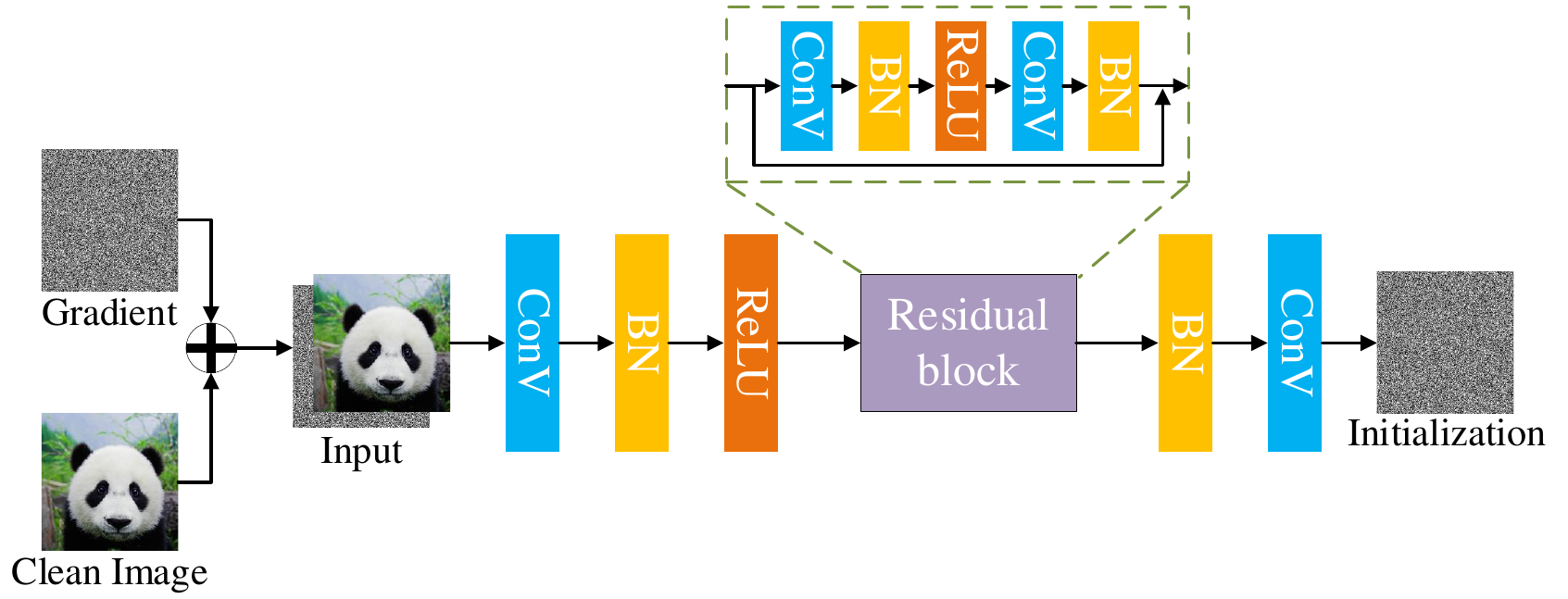}
\end{center}
\caption{The architecture  of our lightweight generative network. The clean image combined with its gradient information from the target network forms the input of the generative network. The  generative network  consists of two convolutional layers and one ResBlock, which outputs the adversarial initialization for the clean image.
}
\label{fig:generative}
\end{figure*}
\begin{table*}[t]
\centering
\caption{The architecture of the generative network.}
\label{tb:structure}
\begin{tabular}{ccccccc}
\hline
Layer     & Type     & Input Channels & Output Channels & Stride & Padding & Filter Size \\ \hline
1st layer & Conv +BN+ ReLU   & 6             & 64              & 1      & 1       & $3 \times 3 $      \\ \hline
2nd layer & ResBlock & 64             & 64              & 1      & 1       & $3 \times 3 $       \\ \hline
3rd layer & Conv +BN     & 64             & 3               & 1      & 1       & $3 \times 3 $        \\ \hline
\end{tabular}
\end{table*}
\subsection{Architecture of the Generative Network} 
\label{sec:architecture}
The architecture of the generative network is shown in Fig.~\ref{fig:generative}. We combine the clean image with its gradient information from the target network to form the input of the generative network. 
The generative network generates a sample-dependent adversarial initialization. We adopt a lightweight generative network, which only includes three layers. The detailed setting of each layer is shown in Table~\ref{tb:structure}. The first layer consists of one convolutional layer with 64 filters of size $3 \times 3 \times 6$ which is followed by a batch normalization layer \cite{ioffe2015batch}. The second layer is a ResBlock \cite{miyato2018spectral} with  64 filters of size $3 \times 3 \times 64$. And the third layer consists of one convolutional layer with 64 filters of size $3 \times 3 \times 3$ which is followed by a batch
normalization layer. We adopt the ReLU \cite{agarap2018deep}  as an activation function. 

The proposed generative network generates a sample-dependent initialization for the clean image based on itself and its gradient information. Compared with the random initialization, the proposed sample-dependent adversarial initialization is more informative. A series of experiments demonstrate that the proposed FGSM-SDI not only overcomes the catastrophic overfitting but also reduces the gap between it and the multi-step AT.

\subsection{Formulation of the Proposed Method} 
\label{sec:formulation}

Similar to the adversarial perturbation of FGSM-RS~\cite{wong2020fast} in Eq.~\ref{eq:rs_perturbation}, our perturbation \textit{i.e.,} the approximate solution of the inner maximization problem, can be written as: 
\begin{align} \label{eq:our_perturbation}
\delta_{g} =  \delta_g(\theta)  &=\Pi_{[-\epsilon, \epsilon]^{d}} [ \eta_{g}+ \alpha \text{sign}(\nabla_{x}\mathcal{L}(f(x +\eta_{g}; \mathbf{w}), y) ) ],
\end{align}
where $\eta_g$ is the adversarial initialization defined in Eq.~\ref{eq:inti}, generated by the generative network.
Note that our perturbation involves the parameters $\theta$ of the generative network via the initialization.
The distinctive difference between the perturbation of FGSM-RS (Eq.~\ref{eq:rs_perturbation}) and that of our FGSM-SDI (Eq.~\ref{eq:our_perturbation}) lies in the initialization term $\eta_g$. 
The initialization of FGSM-RS is randomly sampled without any guidance of other information. Though it can improve the diversity of adversarial examples and the robustness, it encounters the catastrophic overfitting issue that the robustness drops sharply in the late training stage (see Fig.~\ref{fig:ovrrfitting}). 
Differently, our initialization $\eta_g$ is a function of the benign image $x$ and its gradient information $s_x$ that provides some informative guidance on the direction of the initial perturbation.  
It not only overcomes the catastrophic overfitting issue but also greatly improves the robustness compared to current fast AT methods, even comparable to PGD-AT. {Please refer to the results of the comparative experiment in Sec.~\ref{sec:sota}.}


With the definition of our perturbation, the objective function of jointly learning the generative network and the target network can be derived as follows. 
From the objective function of standard AT in Eq.~\ref{Eq:sat}, our solution of the inner maximization problem involves the parameters of the generative network.
When fixing the parameters $\theta$, the solution is approximated by $\delta_g$ in Eq.~\ref{eq:our_perturbation}. 
We can further maximize the loss by searching for better parameters $\theta$, \textit{i.e.,}  $\max_{\theta}~\mathcal{L}(f(x + \delta_g(\theta); \mathbf{w}), y) $. 
Hence, the objective function of our joint optimization can be defined as: 
 \begin{align} \label{eq:game}
\min_{\mathbf{w}} \max_{\theta} \mathbb{E}_{(x, y)\sim \mathcal{D}}  & ~\mathcal{L}(f(x + \delta_g(\theta); \mathbf{w}), y). 
\end{align}

As viewed in Eq.~\ref{eq:game}, the generative network plays a game with the target network. The former maximizes the loss to generate an effective initialization for the adversarial example generation, while the latter minimizes the loss to update the parameters to gain model robustness against adversarial examples. 
More importantly, the generative network can generate initializations according to the robustness of the target model at different training stages. 
This minimax problem can be solved by alternatively optimizing $\mathbf{w}$ and $\theta$. 
Note that we update $\theta$ and $\mathbf{w}$ iteratively. We update $\theta$ every k times of updating $\mathbf{w}$.
And  $k$ is a  hyper-parameter that needs to be tuned.
The algorithm for solving this problem is shown in Algorithm \ref{alg:FGSM_SDI_AT}. 

Eq.~\ref{Eq:sat} is the objective of standard and fast AT methods. Compared to Eq.~\ref{Eq:sat}, our formulation has the following differences. 
First, $\delta$ in Eq.~\ref{Eq:sat} is a variable to optimize, while we replace it with the approximate solution $\delta_g(\theta)$ of the inner maximization. 
$\delta_g(\theta)$ is regarded as a function that involves the parameters of the generative network.
Second, we apply an additional maximization to the parameters of the generative network to further maximize the loss, which forces the competition between the two networks.  

\begin{algorithm}[t]
\caption{FGSM-SDI (Ours)}
\label{alg:FGSM_SDI_AT}
\begin{algorithmic}[1] 
\REQUIRE
The epoch $N$, the maximal perturbation $\epsilon$, the step size $\alpha$, the dataset $\mathcal{D}$ including the benign sample $x$ and the  corresponding label $y$, the dataset size $M$, the target network $f(\cdot, \mathbf{w})$ with parameters $\mathbf{w}$, the generative network $g(\cdot, \theta)$ with parameters $\theta$ and the interval k. 

\FOR{$n=1,...,N$}
\FOR{$i=1,...,M$}

\STATE $ s_{x_{i}}=\text{sign}(\nabla_{x_{i}}\mathcal{L}(f(x_{i}; \mathbf{w}), y_{i}) ) $
\IF{$i  \bmod \text{k} = 0$}
\STATE $ \eta_{g} = \epsilon  g(x_{i},s_{x_{i}};\theta) $
\STATE $ \delta =\Pi_{[-\epsilon, \epsilon]^{d}} [ \eta_{g}+ \alpha \text{sign}(\nabla_{x}\mathcal{L}(f(x_{i} +\eta_{g}; \mathbf{w}), y) ) ]$
\STATE $ \theta \leftarrow \theta +\nabla_{\theta}\mathcal{L}(f(x_{i} + \delta;\theta), y_{i})  $
\ENDIF
\STATE $ \eta_{g} = \epsilon  g(x_i,s_{x_i};\theta) $
\STATE $ \delta =\Pi_{[-\epsilon, \epsilon]^{d}} [ \eta_{g}+ \alpha \text{sign}(\nabla_{x}\mathcal{L}(f(x_{i} +\eta_{g}; \mathbf{w}), y) ) ]$
\STATE $ \mathbf{w} \leftarrow \mathbf{w} -\nabla_{\mathbf{w}}\mathcal{L}(f(x_{i} + \delta; \mathbf{w}), y_{i})  $

\ENDFOR
\ENDFOR
\end{algorithmic}
\end{algorithm}

\textbf{Connection to Two-step PGD-AT.} In our adversarial example generation process (see Fig.~\ref{fig:pipeline}), we calculate the gradient twice with FGSM, \textit{i.e.,} one as input of the generative network for initialization generation and the other for adversarial example generation. 
However, our method is quite different from the two-step PGD-AT method (PGD2-AT) with the number of iterations being 2. 
PGD2-AT can be regarded as a fast AT method that straightforwardly uses the gradient in the first step as initialization. 
Such initialization limits the diversity of adversarial examples as it is bounded by a fixed step size, a pre-defined projection range, and the sign operation (see Eq.~\ref{eq:pgd}). 
Our method uses a generative network to produce the initialization without the setting of step size or projection. The adversarial initialization provides a perturbation to the gradient, which enriches the diversity of adversarial examples and further improves model robustness.
Experimental evaluations show the superiority of our method against PGD2-AT (see Table~\ref{table:ablation_pdg_2}). 

\section{Experiments}
To evaluate the effectiveness of our FGSM-SDI, extensive experiments are conducted on four benchmark databases, including the selection of hyper-parameters in the proposed FGSM-SDI, the ablation study of the adversarial example generation, and the comparisons with state-of-the-art fast AT methods.  

\subsection{Experimental Settings}
\noindent
\textbf{Datasets.} We adopt four benchmark databases to conduct experiments, \textit{i.e.,} CIFAR10 \cite{krizhevsky2009learning}, CIFAR100 \cite{krizhevsky2009learning}, Tiny ImageNet \cite{deng2009ImageNet} and ImageNet \cite{deng2009ImageNet}.
They are the most widely used databases to evaluate adversarial robustness. Both CIFAR10 and CIFAR100 consist of 50,000 training images and 10,000 test images. The image size is $32 \times 32 \times 3$.  
CIFAR10 covers 10 classes while CIFAR100 covers 100 classes.  
Tiny ImageNet is a subset of the ImageNet database \cite{deng2009ImageNet}, which contains 200 classes. Each class has 600 images. The image size is $64 \times 64 \times 3$. As for the ImageNet database, it contains 1000 classes and we resize the image to $224 \times 224 \times 3$.
Following the setting of \cite{lee2020adversarial}, as Tiny ImageNet and ImageNet have no labels for the test dataset, we conduct evaluations on the validation dataset. 

\vspace{1mm}

\label{sec:Experiments Setting}
\noindent
\textbf{Experimental Setups.}  On  CIFAR10, ResNet18 \cite{he2016deep} and WideResNet34-10 \cite{zagoruyko2016wide} are used as the target network. On CIFAR100, ResNet18 \cite{he2016deep} is used as the target network. On Tiny ImageNet, PreActResNet18 \cite{he2016identity} is used as the target network. 
On ImageNet, ResNet50 \cite{he2016deep} is used as the target network.
As for CIFAR10, CIFAR100, and Tiny ImageNet,
following the settings of \cite{rice2020overfitting, pang2020bag}, the target network is trained for 110 epochs. 
The learning rate decays with a factor of 0.1 at the 100-th and 105-th epoch. 
 We adopt the SGD \cite{qian1999momentum} momentum optimizer with an initial learning rate of 0.1 and the weight decay of 5e-4. 
As for ImageNet, following the previous study \cite{shafahi2019adversarial, wong2020fast}, the target network is trained for 90 epochs. The learning rate decays with a factor of 0.1 at the 30-th and 60-th epoch. The SGD momentum optimizer is used with an initial learning rate of 0.1 and the weight decay of 5e-4.
Note that we report the results of the checkpoint with the best accuracy under the attack of PGD-10 as well as the results of the last checkpoint. 
For adversarial robustness evaluation, we adopt several adversarial attack methods to attack the target network, including FGSM~\cite{goodfellow2014explaining}, PGD~\cite{madry2017towards}, C\&W~\cite{carlini2017towards}, and AA \cite{croce2020reliable}. 
And the maximum perturbation strength $\epsilon$ is set to 8/255 for all attack methods.
Moreover, we conduct the PGD attack with 10, 20, and 50 iterations, \textit{i.e.,} PGD-10, PGD-20, and PGD-50. 
We run all our experiments on a single NVIDIA Tesla V100 based on which the training time is calculated. 
We also conduct comparison experiments using a cyclic learning rate strategy \cite{smith2017cyclical}. 

\begin{figure}
\begin{center}
 \includegraphics[width=0.9\linewidth]{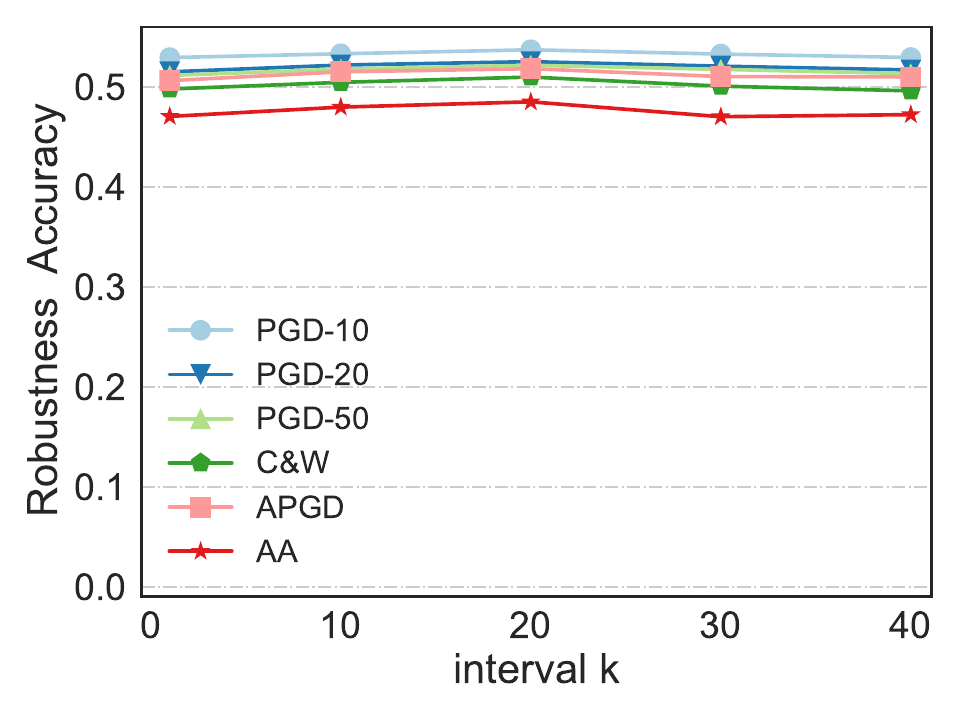}
\end{center}
\vspace{-5mm}
\caption{ Robustness accuracy of the proposed FGSM-SDI with different interval k. We adopt Resnet18 on the CIFAR10 database to conduct experiments}
\label{fig:interval}
\vspace{-5mm}
\end{figure}
\subsection{Hyper-parameter Selection}
\label{selection}
There is one hyper-parameter in the proposed FGSM-SDI, \textit{i.e,} the interval  $\text{k}$. We update $\theta$ every k times of updating $\mathbf{w}$. This hyper-parameter not only affects model training efficiency but also affects model robustness against adversarial examples. To select the optimal hyper-parameter, we conduct a hyper-parameter selection experiment on CIFAR10.
The results are shown in Fig.~\ref{fig:interval}.
The calculation time of the proposed FGSM-SDI decreases along with the increase of parameter $\text{k}$. That is, the smaller the $\text{k}$ value is, the more frequently the generative network is updated, then the generative network requires more calculation time for training. Surprisingly, when $\text{k}= 1 \sim 20 $, the performance against adversarial examples improves with the increase of parameter $\text{k}$. When $\text{k}= 20 \sim 40 $, the performance against adversarial examples slightly drops with the increase of parameter $\text{k}$. When $\text{k}=20$, the proposed FGSM-SDI achieves the best adversarial robustness in all adversarial attack scenarios. Considering adversarial training efficiency, we set $\text{k}$ to 20.

\begin{figure*}
\begin{center}
 \includegraphics[width=0.85\linewidth]{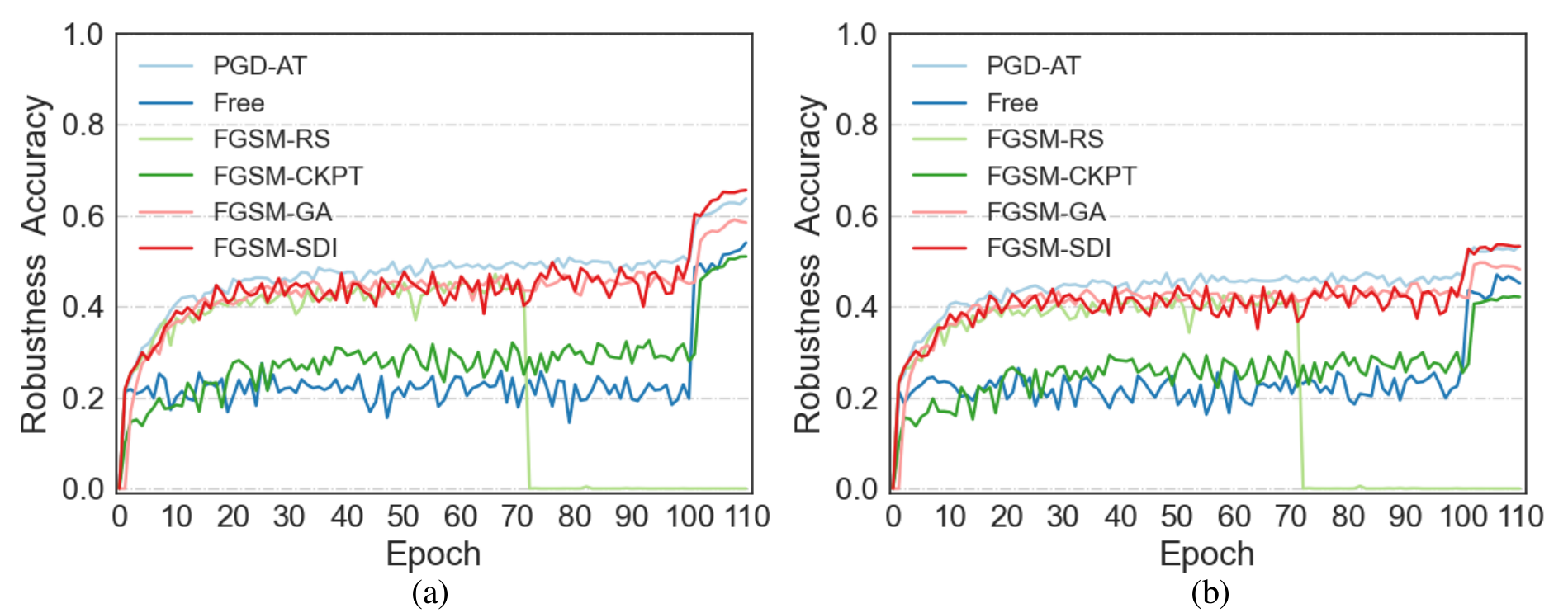}
\end{center}
\vspace{-5mm}
\caption{The PGD-10 accuracy of AT methods on the CIFAR10 database in the training phase. (a) The PGD-10 accuracy on the training dataset. (b) The PGD-10 accuracy on the test dataset. }
\label{fig:ovrrfitting}

\end{figure*}
\begin{table*}[t]
\centering
\caption{Ablation study of the inputs of the generative network on the CIFAR10 database. Numbers in table represent percentage. Number in bold indicates the best.}
 \label{table:ablation}
 \scalebox{1.}{
\begin{tabular}{c|c|c|c|c|c|c|c}
\hline
Input                        &      & Clean          & PGD-10         & PGD-20         & PGD-50         & CW             & AA             \\ \hline
\multirow{2}{*}{Benign}      & Best & 73.34          & 42.63          & 41.82          & 41.66          & 42.31          & 36.72          \\ \cline{2-8} 
                             & Last & \textbf{89.64} & 21.34          & 13.72          & 7.59           & 4.04           & 0.00           \\ \hline
\multirow{2}{*}{Grad}        & Best & \textbf{86.08} & 50.09          & 48.44          & 47.97          & 48.49          & 44.26          \\ \cline{2-8} 
                             & Last & 86.08          & 50.09          & 48.44          & 47.97          & 48.49          & 44.26          \\ \hline
\multirow{2}{*}{Benign+Grad} & Best & 84.86          & \textbf{53.73} & \textbf{52.54} & \textbf{52.18} & \textbf{51.00} & \textbf{48.52} \\ \cline{2-8} 
                             & Last & 85.25          & \textbf{53.18} & \textbf{52.05} & \textbf{51.79} & \textbf{50.29} & \textbf{47.91} \\ \hline
\end{tabular}
}
\end{table*}
\subsection{Ablation Study}
As shown in Fig.~\ref{fig:pipeline}, the generative network takes a benign image and its signed gradient as input to produce an initialization for FGSM to generate adversarial examples. 
Here, we study the influence of each input on the quality of the initialization. Moreover, as mentioned in Sec.~\ref{sec:formulation}, since our method has a certain connection to the two-step PGD-AT, we also conduct experiments to compare with it. 

The results of studying the inputs of the generative network are shown in Table {~\ref{table:ablation}}. `\texttt{Benign}' represents that only the benign image is fed into the generative network, while `\texttt{Grad}' represents that the signed gradient is the only input. `\texttt{Benign} + \texttt{Grad}' represents that both the benign image and signed gradient are regarded as input. 
Analyses are summarized as follows. 
First, it can be observed that \texttt{Benign} + \texttt{Grad} achieves the best performance in robustness under all attacks by exploiting both types of inputs, which indicates that each input contributes to the final performance. 
Second, the gradient information plays a more important role than the benign image because \texttt{Grad} outperforms \texttt{Benign} by a large margin, especially under the attack methods in AA. 
Third, only using a benign image as input cannot produce a desirable initialization and still suffers from the catastrophic overfitting issue, as the robustness of \texttt{Benign} drops dramatically in the late training phase, \textit{i.e.,} the performance of the last checkpoint is much worse than that of the best checkpoint.

The comparison with the two-step PGD-AT (\textit{i.e.,} PGD2-AT~\cite{rice2020overfitting} ) is shown in Table~\ref{table:ablation_pdg_2}. 
Following the setting of \cite{kim2020understanding,andriushchenko2020understanding}, we use the step size $\alpha = \epsilon /2 $ to conduct PGD2-AT. Both PGD2-AT and our FGSM-SDI calculate the gradient w.r.t input sample twice by FGSM to generate adversarial examples. Their differences are stated in Sec.~\ref{sec:formulation}. 
It can be observed that our FGSM-SDI can achieve much better performance in robustness than PGD2-AT in all the attack scenarios. 
For example, compared with PGD2-AT, our FGSM-SDI improves the performance under the strong attack (AA) by about 4\% on the best and last checkpoints.
PGD2-AT can be viewed as a method that uses the gradient of the first FGSM as initialization for the second FGSM. Since the initialization is exactly determined by the gradient, it limits the diversity of subsequently generated adversarial examples. We alleviate this issue by introducing a generative network to learn how to generate initialization under the game mechanism. 
The superiority of our method demonstrates the feasibility of this strategy. Compared with PGD2-AT, our FGSM-SDI costs a little more calculating time, but it achieves much higher robustness under all the attack scenarios. To further verify the effectiveness of our FGSM-SDI, we compare our FGSM-SDI with the four 
PGD-AT (\textit{i.e.,} PGD4-AT~\cite{rice2020overfitting} ). Following the setting of \cite{zhang2020robust}, we use the step size $\alpha = \epsilon /4 $ to conduct PGD4-AT. The result is shown in Table ~\ref{table:ablation_pdg_2}. Compared with PGD4-AT, our FGSM-SDI not only achieves much better robustness performance in all the attack scenarios but also costs less calculating time for training. Specifically, as for the strong attack (AA), our FGSM-SDI improves the performance by about 3\%  on the best and last checkpoints. As for the training efficiency, our FGSM-SDI reduces the training time by about 36 minutes.
Note that there is a trade-off between the clean accuracy and the robustness, better robustness always comes along with a lower clean accuracy~\cite{zhang2019theoretically}. Although our FGSM-SDI reduces the clean accuracy slightly, it improves the robust accuracy greatly.

\begin{table*}[t]
\centering
\caption{Comparisons with PGD2-AT and PGD4-AT on CIFAR10 database. Numbers in table represent percentage. Number in bold indicates the best. }
 \label{table:ablation_pdg_2}
\scalebox{0.95}{
\begin{tabular}{c|c|c|c|c|c|c|c|c|c}
\hline
Method                          &      & Clean & PGD-10         & PGD-20         & PGD-50         & C\&W             & APGD           & AA             & Time(min)            \\ \hline
\multirow{2}{*}{PGD2-AT}        & Best & \textbf{86.28} & 49.28          & 47.51          & 47.01          & 47.73          & 46.56          & 44.47          & \multirow{2}{*}{77}  \\ \cline{2-9}
                                & Last & \textbf{86.64} & 48.49          & 47.05          & 46.46          & 47.31          & 45.98          & 44.14          &                      \\ \hline
\multirow{2}{*}{PGD4-AT}        & Best & 86.15 & 49.44          & 48.08          & 47.56          & 48.11          & 47.22          & 45.11          & \multirow{2}{*}{119} \\ \cline{2-9}
                                & Last & 86.61 & 48.94          & 47.27          & 46.88          & 47.82          & 46.63          & 44.60          &                      \\ \hline
\multirow{2}{*}{FGSM-SDI(ours)} & Best & 84.86 & \textbf{53.73} & \textbf{52.54} & \textbf{52.18} & \textbf{51.00} & \textbf{51.84} & \textbf{48.50} & \multirow{2}{*}{83}  \\ \cline{2-9}
                                & Last & 85.25 & \textbf{53.18} & \textbf{52.05} & \textbf{51.79} & \textbf{50.29} & \textbf{51.30} & \textbf{47.91} &                      \\ \hline
\end{tabular}
}
\end{table*}


\subsection{Relieving Catastrophic Overfitting}
Catastrophic overfitting {~\cite{wong2020fast,andriushchenko2020understanding,kim2020understanding}} is one of the tough problems that mostly affect the model robustness of the fast AT methods, which refers to the phenomenon that the accuracy on adversarial examples suddenly drops to 0.00. 
To investigate the catastrophic overfitting, we record the accuracy of adversarial examples generated on training and test data in the training phase. Adversarial examples are generated by PGD-10. 
The training and test curves under the PGD-10 attack are shown in Fig. \ref{fig:ovrrfitting}. The accuracy of FGSM-RS  decreases rapidly to 0 after about 70-th epoch.
Although other fast AT methods overcome the catastrophic overfitting issue, their performance is far from satisfactory, \textit{i.e.,} there is a huge gap between the fast AT methods and the advanced multi-step PGD-AT \cite{rice2020overfitting}. 
For example, FGSM-GA achieves the accuracy of 58.46\% and 48.17\%  on training and test data respectively, while PGD-AT achieves the much higher accuracy of 63.71\% and 53.33\%. 
Differently, our FGSM-SDI can not only overcome the catastrophic overfitting but also achieve comparable robustness to PGD-AT. 
Specifically, it achieves the accuracy of 64.14\% and 52.81\% on training and test data, respectively. 
Note that our FGSM-SDI costs only one-third the computation time of PGD-AT and also less time than FGSM-GA. More details will be discussed in the following part.
\begin{table*}[t]
\caption{Comparisons of clean and robust accuracy (\%) and training time (minute) with Resnet18 on the CIFAR10 database. Number in bold indicates the best of the fast AT methods. }
\centering
\label{table:cifar10}
 \scalebox{0.9}{
\begin{tabular}{c|c|c|c|c|c|c|c|c|c|c}
\hline
Target Network             & Method                          &      & Clean          & PGD-10         & PGD-20         & PGD-50         & C\&W             & APGD           & AA             & Time(min)            \\ \hline
\multirow{2}{*}{Resnet18}  & \multirow{2}{*}{PGD-AT}         & Best & 82.32          & 53.76          & 52.83          & 52.6         & 51.08          & 52.29          & 48.68          & \multirow{2}{*}{265} \\ \cline{3-10}
                           &                                 & Last & 82.65          & 53.39          & 52.52          & 52.27          & 51.28          & 51.90          & 48.93          &                      \\ \hline \hline
\multirow{10}{*}{Resnet18} & \multirow{2}{*}{FGSM-RS}        & Best & 73.81          & 42.31          & 41.55          & 41.26          & 39.84          & 41.02          & 37.07          & \multirow{2}{*}{51}  \\ \cline{3-10}
                           &                                 & Last & 83.82          & 00.09          & 00.04          & 00.02          & 0.00           & 0.00           & 0.00           &                      \\ \cline{2-11} 
                           & \multirow{2}{*}{FGSM-CKPT}      & Best & \textbf{90.29} & 41.96          & 39.84         & 39.15          & 41.13          & 38.45          & 37.15          & \multirow{2}{*}{76}  \\ \cline{3-10}
                           &                                 & Last & \textbf{90.29} & 41.96          & 39.84         & 39.15          & 41.13          & 38.45          & 37.15         &                      \\ \cline{2-11} 
                           & \multirow{2}{*}{FGSM-GA}        & Best & 83.96          & 49.23          & 47.57          & 46.89          & 47.46          & 45.86          & 43.45          & \multirow{2}{*}{178} \\ \cline{3-10}
                           &                                 & Last & 84.43          & 48.67          & 46.66          & 46.08          & 46.75          & 45.05          & 42.63          &                      \\ \cline{2-11} 
                           & \multirow{2}{*}{Free-AT(m=8)}   & Best & 80.38          & 47.1           & 45.85          & 45.62           & 44.42          & 42.18          & 42.17          & \multirow{2}{*}{215} \\ \cline{3-10}
                           &                                 & Last & 80.75          & 45.82          & 44.82         & 44.48          & 43.73          & 45.22          & 41.17         &                      \\ \cline{2-11} 
                           & \multirow{2}{*}{FGSM-SDI(ours)} & Best & 84.86          & \textbf{53.73} & \textbf{52.54} & \textbf{52.18} & \textbf{51.00} & \textbf{51.84} & \textbf{48.50} & \multirow{2}{*}{83}  \\ \cline{3-10}
                           &                                 & Last & 85.25          & \textbf{53.18} & \textbf{52.05} & \textbf{51.79} & \textbf{50.29} & \textbf{51.30} & \textbf{47.91} &                      \\ \hline
\end{tabular}
}
\end{table*}
\begin{table*}[t]
\caption{Comparisons of clean and robust accuracy (\%) and training time (minute) with WideResNet34-10 on the CIFAR10 database. Number in bold indicates the best of the fast AT methods. }
\centering
\label{table:cifar10_wide}
 \scalebox{0.9}{
\begin{tabular}{c|c|c|c|c|c|c|c|c|c}
\hline
Target Network                   & Method         & Clean          & PGD-10         & PGD-20         & PGD-50        & C\&W           & APGD           & AA             & Time(min) \\ \hline
WideResNet34-10                  & PGD-AT         & 85.17          & 56.1           & 55.07          & 54.87         & 53.84          & 54.15          & 51.67          & 1914      \\ \hline \hline
\multirow{5}{*}{WideResNet34-10} & FGSM-RS        & 74.29          & 41.24          & 40.21          & 39.98         & 39.27          & 39.79          & 36.40          & 348       \\ \cline{2-10} 
                                 & FGSM-CKPT      & \textbf{91.84} & 44.7           & 42.72          & 42.22         & 42.25          & 41.69          & 40.46          & 470       \\ \cline{2-10} 
                                 & FGSM-GA        & 81.8           & 48.2           & 47.97          & 46.6          & 46.87          & 46.27          & 45.19          & 1218      \\ \cline{2-10} 
                                 & Free-AT(m=8)   & 81.83          & 49.07          & 48.17          & 47.83         & 47.25          & 47.40          & 44.77          & 1422      \\ \cline{2-10} 
                                 & FGSM-SDI(ours) & 86.4  & \textbf{55.89} & \textbf{54.95} & \textbf{54.6} & \textbf{53.68} & \textbf{54.21} & \textbf{51.17} & 533       \\ \hline
\end{tabular}
}
\end{table*}
\begin{table*}[t]
\caption{Comparisons of clean and robust accuracy (\%) and training time (minute) on the CIFAR10 database. Number in bold indicates the best of the fast AT methods. \textbf{All models are trained using a
cyclic learning rate strategy}. }
\centering
\label{table:cifar10_cyclic}
 \scalebox{0.9}{
\begin{tabular}{c|c|c|c|c|c|c|c|c|c|c}
\hline
Target Network             & Method                          &      & Clean          & PGD-10         & PGD-20         & PGD-50         & CW             & APGD           & AA             & Time(min)           \\ \hline 
\multirow{2}{*}{Resnet18}  & \multirow{2}{*}{PGD-AT}         & Best & 80.12          & 51.59          & 50.83          & 50.7           & 49.04          & 50.34          & 46.83          & \multirow{2}{*}{71} \\ \cline{3-10}
                           &                                 & Last & 80.12          & 51.59          & 50.83          & 50.7           & 49.04          & 50.34          & 46.83          &                     \\ \hline  \hline
\multirow{10}{*}{Resnet18} & \multirow{2}{*}{FGSM-RS}        & Best & 83.75          & 48.05          & 46.47          & 46.11          & 46.21          & 45.75          & 42.92          & \multirow{2}{*}{15} \\ \cline{3-10}
                           &                                 & Last & 83.75          & 48.05          & 46.47          & 46.11          & 46.21          & 45.75          & 42.92          &                     \\ \cline{2-11} 
                           & \multirow{2}{*}{FGSM-CKPT}      & Best & \textbf{89.08} & 40.47          & 38.2           & 37.69          & 39.87          & 37.16          & 35.81          & \multirow{2}{*}{21} \\ \cline{3-10}
                           &                                 & Last & \textbf{89.08} & 40.47          & 38.2           & 37.69          & 39.87          & 37.16          & 35.81          &                     \\ \cline{2-11} 
                           & \multirow{2}{*}{FGSM-GA}        & Best & 80.83          & 48.76          & 47.83          & 47.54          & 47.14          & 47.27          & 44.06          & \multirow{2}{*}{49} \\ \cline{3-10}
                           &                                 & Last & 80.83          & 48.76          & 47.83          & 47.54          & 47.14          & 47.27          & 44.06          &                     \\ \cline{2-11} 
                           & \multirow{2}{*}{Free-AT(m=8)}   & Best & 75.22          & 44.67          & 43.97          & 43.72          & 42.48          & 43.55          & 40.30          & \multirow{2}{*}{59} \\ \cline{3-10}
                           &                                 & Last & 75.22          & 44.67          & 43.97          & 43.72          & 42.48          & 43.55          & 40.30          &                     \\ \cline{2-11} 
                           & \multirow{2}{*}{FGSM-SDI(ours)} & Best & 82.08          & \textbf{51.63} & \textbf{50.65} & \textbf{50.33} & \textbf{48.57} & \textbf{49.98} & \textbf{46.21} & \multirow{2}{*}{23} \\ \cline{3-10}
                           &                                 & Last & 82.08          & \textbf{51.63} & \textbf{50.65} & \textbf{50.33} & \textbf{48.57} & \textbf{49.98} & \textbf{46.21} &                     \\ \hline
\end{tabular}
}
\end{table*}
\begin{table*}[t]
\centering
\caption{Comparisons of clean and robust accuracy (\%) and training time (minute) with Resnet18 on the CIFAR100 database. Number in bold indicates the best of the fast AT methods. }
\label{table:cifar100}
 \scalebox{0.9}{
\begin{tabular}{c|c|c|c|c|c|c|c|c|c|c}
\hline
Target Network             & Method                          &      & Clean          & PGD-10         & PGD-20         & PGD-50         & C\&W             & APGD           & AA             & Time(min)            \\ \hline
\multirow{2}{*}{Resnet18}  & \multirow{2}{*}{PGD-AT}         & Best & 57.52          & 29.6           & 28.99          & 28.87          & 28.85          & 28.60          & 25.48          & \multirow{2}{*}{284} \\ \cline{3-10}
                           &                                 & Last & 57.5           & 29.54          & 29.00          & 28.90          & 27.6           & 28.70          & 25.48          &                      \\ \hline \hline
\multirow{10}{*}{Resnet18} & \multirow{2}{*}{FGSM-RS}        & Best & 49.85          & 22.47          & 22.01          & 21.82          & 20.55          & 21.62          & 18.29          & \multirow{2}{*}{70}  \\ \cline{3-10}
                           &                                 & Last & 60.55          & 00.45          & 00.25          & 00.19          & 00.25          & 0.00           & 0.00           &                      \\ \cline{2-11} 
                           & \multirow{2}{*}{FGSM-CKPT}      & Best & \textbf{60.93} & 16.58          & 15.47          & 15.19          & 16.4           & 14.63          & 14.17          & \multirow{2}{*}{96}  \\ \cline{3-10}
                           &                                 & Last & \textbf{60.93} & 16.69          & 15.61          & 15.24          & 16.6           & 14.87          & 14.34          &                      \\ \cline{2-11} 
                           & \multirow{2}{*}{FGSM-GA}        & Best & 54.35          & 22.93          & 22.36           & 22.2         & 21.2          & 21.88          &     18.88      & \multirow{2}{*}{187} \\ \cline{3-10}
                           &                                 & Last & 55.1           & 20.04          & 19.13          & 18.84          & 18.96          & 18.46          & 16.45          &                      \\ \cline{2-11} 
                           & \multirow{2}{*}{Free-AT(m=8)}   & Best & 52.49          & 24.07          & 23.52         & 23.36          & 21.66          & 23.07          & 19.47         & \multirow{2}{*}{229} \\ \cline{3-10}
                           &                                 & Last & 52.63          & 22.86          & 22.32          & 22.16          & 20.68         &  21.90          & 18.57          &                      \\ \cline{2-11} 
                           & \multirow{2}{*}{FGSM-SDI(ours)} & Best & 60.67          & \textbf{31.5}  & \textbf{30.89} & \textbf{30.6}  & \textbf{27.15} & \textbf{30.26} & \textbf{25.23} & \multirow{2}{*}{99}  \\ \cline{3-10}
                           &                                 & Last & 60.82          & \textbf{30.87} & \textbf{30.34} & \textbf{30.08} & \textbf{27.3}  & \textbf{29.94} & \textbf{25.19} &                      \\ \hline
\end{tabular}
}
\end{table*}

\subsection{Comparisons with State-of-the-art Methods} \label{sec:sota}
We compare our FGSM-SDI with several state-of-the-art fast AT methods (\textit{i.e.,} Free \cite{shafahi2019adversarial}, FGSM-RS \cite{wong2020fast}, FGSM-GA \cite{andriushchenko2020square}, and FGSM-CKPT \cite{kim2020understanding}) and an advanced multi-step AT method ( \textit{i.e.,} PGD-AT \cite{rice2020overfitting}) which adopts 10 steps to generate adversarial examples on four benchmark databases. We follow the settings reported in their original works to train these AT methods. Note that to ensure fairness of comparison, we do not divide the number of epochs by $m$ such that the total number of epochs remains the same as the other fast AT methods.
\begin{figure}
\begin{center}
 \includegraphics[width=0.9\linewidth]{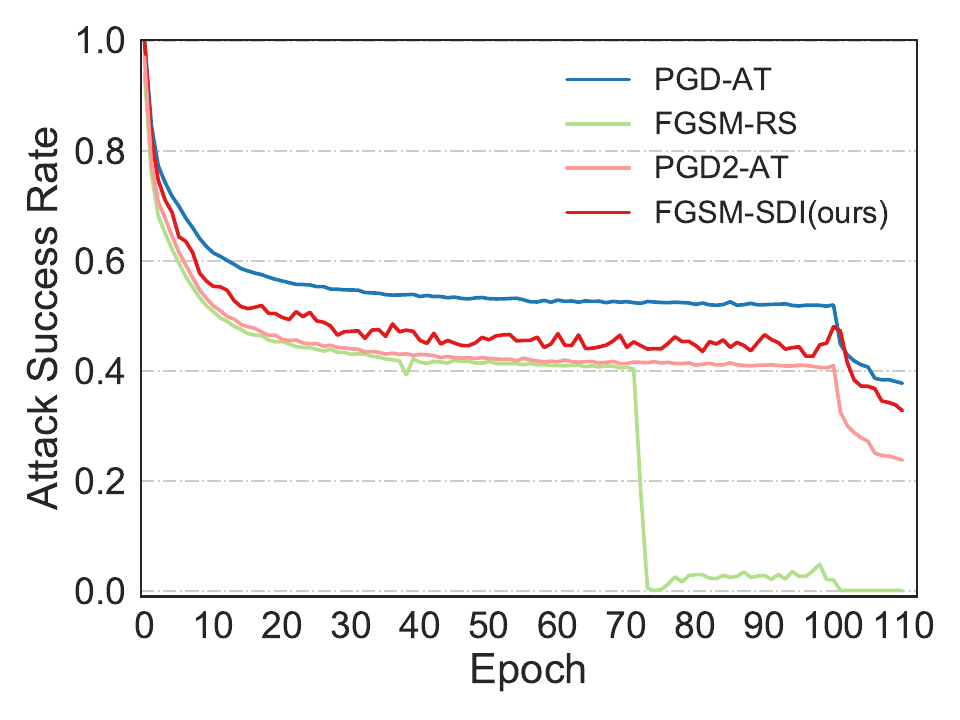}
\end{center}
\vspace{-5mm}
\caption{ Attack success rate of FGSM-RS, PGD-AT, PGD2-AT and FGSM-SDI(ours) during the training process.}
\label{fig:diff}
\vspace{-5mm}
\end{figure}
\begin{figure}
\begin{center}
 \includegraphics[width=0.9\linewidth]{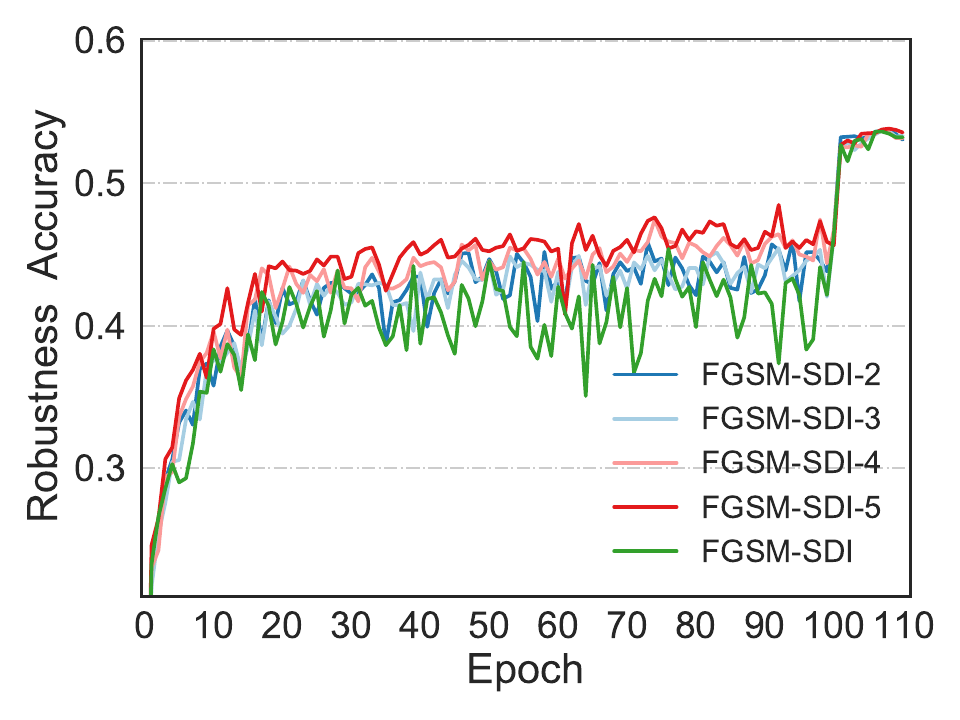}
\end{center}
\vspace{-5mm}
\caption{ The PGD-10 accuracy of FGSM-SDI with different m iterations of the generate network on the CIFAR10 database in the training phase.}
\label{fig:overfitting_iter}
\vspace{-5mm}
\end{figure}
\begin{figure}
\begin{center}
 \includegraphics[width=1\linewidth]{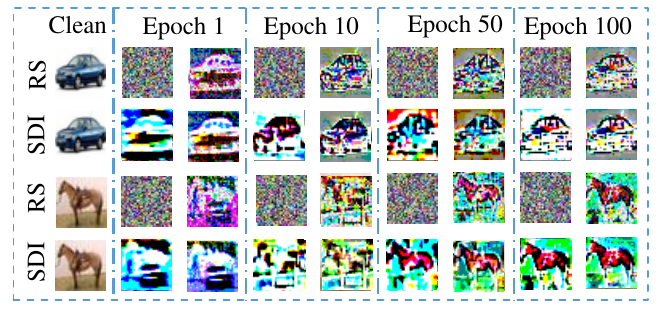}
\end{center}
\vspace{-5mm}
\caption{ Visualization of the adversarial initialization and FGSM-updated perturbations for the FGSM-RS and FGSM-SDI among continuous training epochs.}
\label{fig:init_diff}
\vspace{-5mm}
\end{figure}
\begin{figure*}
\begin{center}
 \includegraphics[width=1\linewidth]{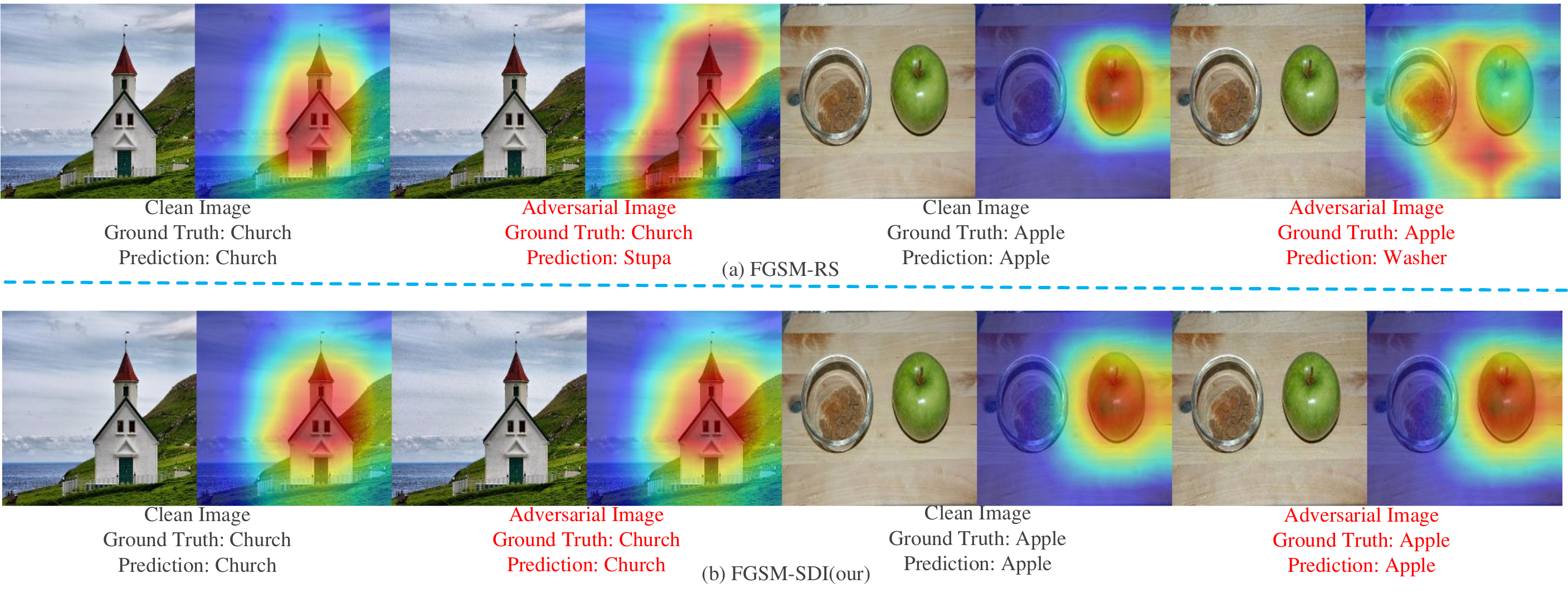}
\end{center}
\vspace{-3mm}
\caption{The top row shows the clean images and the adversarial examples along with their corresponding heat-maps (generated by the Grad-CAM algorithm) on the FGSM-RS. The bottom row shows the results of our FGSM-SDI. {Note that we adopt the same adversarial attack \textit{i.e.,} PGD-10 , to conduct the visualization.} }
\label{fig:cam}
\vspace{-3mm}
\end{figure*}
\begin{figure*}
\begin{center}
 \includegraphics[width=1\linewidth]{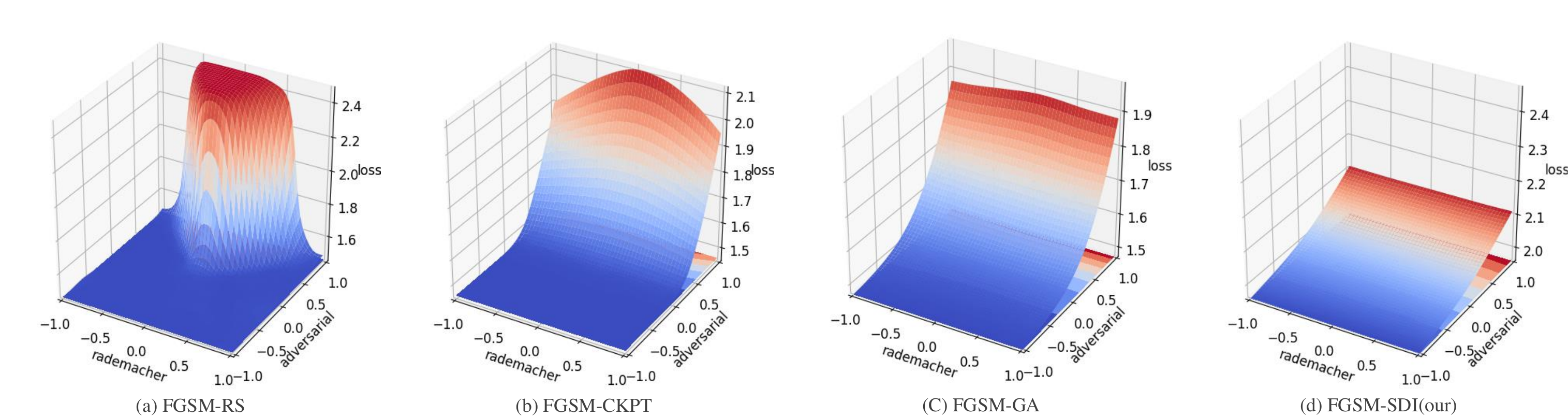}
\end{center}
\vspace{-3mm}
\caption{Visualization of the loss landscape of on CIFAR10 for FGSM-RS, FGSM-CKPT, FGSM-GA, and our FGSM-SDI. We plot the cross entropy loss varying along the space consisting of two directions: an adversarial direction $r_{1}$ and a Rademacher (random) direction $r_{2}$. The adversarial direction can be defined as: $r_{1}=\eta \operatorname{sign}(\nabla_{x} f(\hat{x}))$ and the Rademacher (random) direction can be defined as: $r_{2} \sim \operatorname{Rademacher}(\eta)$, where $\eta$ is set to $8/255$. {Note that we adopt the same adversarial attack \textit{i.e.,} PGD-10 , to conduct the visualization.} }
\label{fig:loss_land}
\vspace{-3mm}
\end{figure*}
\begin{figure}
\begin{center}
 \includegraphics[width=0.8\linewidth]{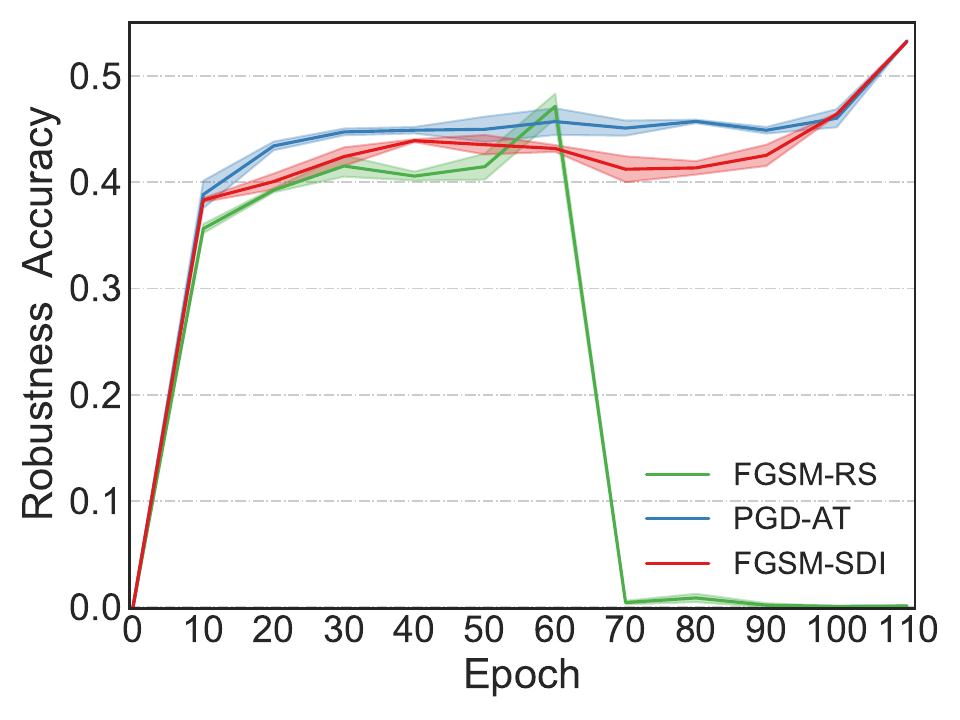}
\end{center}
\vspace{-5mm}
\caption{ The PGD-10 accuracy of FGSM-RS, PGD-AT and our FGSM-SDI with multiple training  on the CIFAR10 database in the training phase.}
\label{fig:mean_std}
\vspace{-5mm}
\end{figure}

\noindent \textbf{Results on CIFAR10.}
We adopt Resnet18 as the target network to conduct the comparison experiment with other defense methods on CIFAR10.
The results are shown in Table \ref{table:cifar10}. 
Compared with the fast AT methods, our method achieves the best performance under all attack scenarios and comparable robustness to the advanced PGD-AT \cite{rice2020overfitting}.
The previous most fast AT methods 
are only as effective as the prime PGD-AT \cite{madry2017towards},  \textit{i.e.,} they achieve the performance of about 45\% under the PGD-10 attack. 
The performance is far from that of the advanced PGD-AT \cite{rice2020overfitting} which uses an early stopping trick to achieve above 50\% accuracy. 
Unlike them, our method can achieve more than 53\%  under the PGD-10 attack on  the  best  and  last  checkpoint. As for the strong attack (AA), the previous most powerful fast AT method (FGSM-GA) achieves the performance of about 43\%, but the proposed FGSM-SDI achieves about 48\% robust accuracy which is the same as PGD-AT.
In terms of training efficiency, our training time is less than FGSM-GA, Free, and PGD-AT. Specifically, the training time of our FGSM-SDI is about  1.6 times the training time of FGSM-RS \cite{wong2020fast}.
Though FGSM-RS and FGSM-CKPT are more efficient than our method, their performance is always the worst among all the fast AT methods. 
FGSM-RS is the fastest method that uses a random initialization. 
Our method improves the initialization to boost the robustness by introducing the generative network, resulting in the sacrifice of efficiency for an additional gradient calculation. 
Therefore, our method can be viewed as a method that balances the robustness and the training efficiency.

\par Moreover, we adopt WideResNet34-10 which is a large architecture model to conduct a comparison experiment. The results are shown in  Table~\ref{table:cifar10_wide}. We observe a similar  phenomenon as the Resnet18 trained on CIFAR10. Our FGSM-SDI achieves the best performance under  all  attack  scenarios compared with previous fast AT methods. Besides, compared with the advanced PGD-AT, the proposed FGSM-SDI costs less time and achieves  comparable  robustness to it. Specifically, PGD-AT achieves the performance of about 51\% and 49\% against AA attack on the best and last  checkpoints. Our FGSM-SDI also achieves the performance of about 51\% and 49\%. But PGD-AT takes about 1914  minutes for training, while our FGSM-SDI only takes about 533 minutes for training.

\par We also conduct comparative experiments using a cyclic learning rate strategy~\cite{smith2017cyclical} on CIFAR10. Following \cite{andriushchenko2020understanding, kim2020understanding}, we set the maximal learning rate of FGSM-GA \cite{andriushchenko2020understanding} and FGSM-CKPT \cite{kim2020understanding} to 0.3. Following \cite{wong2020fast}, we set the maximal  learning rate of FGSM-RS \cite{wong2020fast}, Free \cite{shafahi2019adversarial}, PGD-AT \cite{rice2020overfitting}, and the proposed method to 0.2.
We train all the models for 30 epochs. Other training and evaluation settings remain unchanged. The results are shown in Table~\ref{table:cifar10_cyclic}.  Compared with the other fast AT methods, the proposed FGSM-SDI achieves the best adversarial robustness and 
comparable robustness to the advanced PGD-AT \cite{rice2020overfitting}. Using a cyclic learning rate strategy can prevent catastrophic overfitting for the fast AT methods,  \textit{i.e.,} the performance of the last checkpoint is almost the same as that of the best checkpoint. But their adversarial robustness is still far from that of the advanced PGD-AT \cite{rice2020overfitting}.
Differently, our FGSM-SDI can achieve comparable robustness to PGD-AT \cite{rice2020overfitting}. For example, FGSM-RS \cite{wong2020fast} 
achieves about 42\% accuracy under AA attack, while our method 
achieves about 46\%. In terms of efficiency, our method outperforms Free and FGSM-GA  and is much better than PGD-AT. FGSM-RS and FGSM-CKPT use a random initialization which promotes efficiency at a sacrifice in robustness.   
Our method improves the initialization with a generative network, which greatly boosts the adversarial robustness with the slight expense of time cost.

\begin{table*}[t]
\centering
\caption{Comparisons of clean and robust accuracy (\%) and training time (minute) with PreActResNet18 on the Tiny ImageNet database. Number in bold indicates the best of the fast AT methods. }
\label{table:Imagnet}
 \scalebox{0.9}{
\begin{tabular}{c|c|c|c|c|c|c|c|c|c|c}
\hline
Target Network                   & Method                          &      & Clean          & PGD-10         & PGD-20         & PGD-50         & CW             & APGD           & AA             & Time(min)             \\ \hline
\multirow{2}{*}{PreActResNet18}  & \multirow{2}{*}{PGD-AT}         & Best & 43.6           & 20.2           & 19.9           & 19.86          & 17.5           & 19.64          & 16.00          & \multirow{2}{*}{1833} \\ \cline{3-10}
                                 &                                 & Last & 45.28          & 16.12          & 15.6           & 15.4           &        14.28        & 15.22          & 12.84          &                       \\ \hline  \hline
\multirow{10}{*}{PreActResNet18} & \multirow{2}{*}{FGSM-RS}        & Best & 44.98          & 17.72          & 17.46          & 17.36          & 15.84          & 17.22          & 14.08          & \multirow{2}{*}{339}  \\ \cline{3-10}
                                 &                                 & Last & 45.18          & 0.00           & 0.00           & 0.00           & 0.00           & 0.00           & 0.00           &                       \\ \cline{2-11} 
                                 & \multirow{2}{*}{FGSM-CKPT}      & Best & \textbf{49.98} & 9.20           & 9.20           & 8.68           & 9.24           & 8.50           & 8.10           & \multirow{2}{*}{464}  \\ \cline{3-10}
                                 &                                 & Last & \textbf{49.98} & 9.20           & 9.20           & 8.68           & 9.24           & 8.50           & 8.10           &                       \\ \cline{2-11} 
                                 & \multirow{2}{*}{FGSM-GA}        & Best & 34.04          & 5.58           & 5.28           & 5.1            & 4.92           & 4.74           & 4.34           & \multirow{2}{*}{1054} \\ \cline{3-10}
                                 &                                 & Last & 34.04          & 5.58           & 5.28           & 5.1            & 4.92           & 4.74           & 4.34           &                       \\ \cline{2-11} 
                                 & \multirow{2}{*}{Free-AT(m=8)}   & Best & 38.9           & 11.62          & 11.24          & 11.02          & 11.00          & 10.88          & 9.28           & \multirow{2}{*}{1375} \\ \cline{3-10}
                                 &                                 & Last & 40.06          & 8.84           & 8.32           & 8.2            & 8.08           & 7.94           & 7.34           &                       \\ \cline{2-11} 
                                 & \multirow{2}{*}{FGSM-SDI(ours)} & Best & 46.46          & \textbf{23.22} & \textbf{22.84} & \textbf{22.76} & \textbf{18.54} & \textbf{22.56} & \textbf{17.00} & \multirow{2}{*}{565}  \\ \cline{3-10}
                                 &                                 & Last &     47.64
           &        \textbf{19.84}        &    \textbf{19.36}           &   \textbf{ 19.16}            &         \textbf{16.02}       &          \textbf{19.08}      &     \textbf{14.10}        &                       \\ \hline
\end{tabular}
}
\vspace{-2mm}
\end{table*}

\noindent \textbf{Results on CIFAR100.}
 The results are shown in Table \ref{table:cifar100}. 
 The CIFAR100 database covers more classes than the CIFAR10, which makes the target network harder to obtain robustness. 
We can observe a similar phenomenon as on CIFAR10. 
In detail, compared with the other fast AT methods, 
our FGSM-SDI achieves the best adversarial robustness under all adversarial attack scenarios.
For example, the previous fast AT methods achieve the performance of about 20\% under the PGD-50 attack which is far from that of the advanced PGD-AT~\cite{rice2020overfitting} which achieves about 28\% accuracy. While the proposed FGSM-SDI achieves the performance of about 30\% under the PGD-50 attack. Surprisingly, our method can even outperform PGD-AT under the attacks of PGD-10, PGD-20, PGD-50, and APGD. Our method also achieves comparable  robustness to the advanced PGD-AT under the strong attack methods (C\&W and AA). And our clean accuracy is also about 3\% higher than PGD-AT~\cite{rice2020overfitting}. This indicates the potential of our method in boosting robustness.  In terms of training efficiency,  similar results are observed on CIFAR10. Our FGSM-SDI can be 3 times faster than the advanced PGD-AT~\cite{rice2020overfitting}. 
Although our FGSM-SDI costs a little more time than FGSM-RS, it not only relieves the catastrophic overfitting problem but also achieves comparable  robustness to the advanced PGD-AT.

\vspace{1mm}
\noindent \textbf{Results on Tiny ImageNet.} 
The results are shown in Table \ref{table:Imagnet}. 
Tiny ImageNet is a larger database compared to CIFAR10 and CIFAR100. 
Performing AT on Tiny ImageNet requires more computational cost. 
It takes about 1,833 minutes for PGD-AT to conduct training. But, our FGSM-SDI only takes 565 minutes and achieves comparable or even better robustness than PGD-AT. Similar to the results on CIFAR100 and CIFAR10, compared with previous fast AT methods, our FGSM-SDI achieves the best performance under all attack scenarios. Moreover, compared with the advanced PGD-AT, our FGSM-SDI achieves better performance under all attack scenarios even the strong attack (AA). Specifically, PGD-AT achieves the performance of about 16\% and 13\% accuracy under AA attack on the best and last checkpoints, while our FGSM-SDI  achieves the performance of about 17\% and 14\% accuracy. Moreover, FGSM-SDI achieves higher clean accuracy compared with PGD-AT. Specifically, our clean accuracy is also about 3\% higher than PGD-AT. The efficiency comparison is similar to that on CIFAR10 and CIFAR100. 
\begin{table}[]
\caption{Comparisons of clean and robust accuracy (\%) and training time (minute) with Resnet50 on the  ImageNet database. Number in bold indicates the best of the fast AT methods. }
\label{table:Larger_Imagnet}

\resizebox{\columnwidth}{!}{

\begin{tabular}{c|c|c|c|c|c}
\hline
ImageNet                         & Epsilon      & Clean  & PGD-10 & PGD-50 & Time(hour)             \\ \hline
\multirow{3}{*}{PGD-AT}          & $\epsilon=$2 &  64.81  & 47.99 &  47.98 & \multirow{3}{*}{211.2} \\ \cline{2-5}
                                 & $\epsilon=$4 & 59.19 & 35.87 & 35.41 &                        \\ \cline{2-5}
                                 & $\epsilon=$8 & 49.52 & 26.19 & 21.17 &                        \\ \hline\hline
\multirow{3}{*}{Free-AT(m=4)}    & $\epsilon=$2 & \textbf{68.37}  & 48.31  & 48.28  & \multirow{3}{*}{127.7} \\ \cline{2-5}
                                 & $\epsilon=$4 & 63.42  & 33.22  & 33.08  &                        \\ \cline{2-5}
                                 & $\epsilon=$8 & 52.09  & 19.46  & 12.92  &                        \\ \hline
\multirow{3}{*}{FGSM-RS}         & $\epsilon=$2 & 67.65  & 48.78  & 48.67  & \multirow{3}{*}{44.5}  \\ \cline{2-5}
                                 & $\epsilon=$4 & \textbf{63.65}  & 35.01  & 32.66  &                        \\ \cline{2-5}
                                 & $\epsilon=$8 & \textbf{53.89}  & 0.00   & 0.00   &                        \\ \hline
\multirow{3}{*}{FGSM-SDI (ours)} & $\epsilon=$2 &     66.01   &    \textbf{49.51}   &    \textbf{49.35}    & \multirow{3}{*}{66.8}  \\ \cline{2-5}
                                 & $\epsilon=$4 &    59.62    &    \textbf{37.5}    &     \textbf{36.63}   &                        \\ \cline{2-5}
                                 & $\epsilon=$8 &   48.51     &   \textbf{26.64}     &   \textbf{21.61}     &                        \\ \hline
\end{tabular}
}
\vspace{-5mm}
\end{table}

\vspace{1mm}

\noindent \textbf{Results on ImageNet.} 
Following \cite{shafahi2019adversarial, wong2020fast}, we adopt Resnet50 to conduct AT on ImageNet under the maximum perturbation strength $\epsilon=2$, $\epsilon=4$, and $\epsilon=8$. The results are shown in Table~\ref{table:Larger_Imagnet}. 
When $\epsilon=2$, all methods achieve roughly the same robustness against adversarial examples. But as the maximal perturbation strength becomes larger, PGD-AT and our FGSM-SDI  achieves better robustness performance. Especially, when $\epsilon=8$, the FGSM-RS cannot defend against the PGD-based attacks. But our FGSM-SDI still achieves the  performance  of  about  26\%  and 21\% under  the  PGD-10 and PGD-50 attacks and achieves  comparable robustness to PGD-AT. In  terms  of  training  efficiency,  similar phenomenons are observed on other databases,  our FGSM-SDI  can  be  3 times  faster  than  the  advanced  PGD-AT.


\subsection{Performance Analysis}
 To explore how our initialization affects the generation of adversarial examples, we train a Renet18 on CIFAR10 and calculate the attack success rate of adversarial examples that successfully attack the target model during the training process. The comparisons with FGSM-RS, PGD2-AT, and PGD-AT are shown in Fig.~\ref{fig:diff}. From the 0-th to 70-th epoch, the attack success rates of the successful adversarial examples of the FGSM-RS, PGD2-AT, and FGSM-SDI are roughly the same. However, after the 70-th epoch, the attack success rate of FGSM-RS drops sharply. At that time the trained model using FGSM-RS falls into the catastrophic overfitting problem that the trained model cannot defend against the adversarial examples generated by PGD-based attack methods during the  training process. While the adversarial examples generated by the other three methods always keep adversarial to
the trained model. They do not meet the catastrophic overfitting. This observation indicates that
the catastrophic overfitting is associated with the adversarial example quality in the training process. Moreover, the attack success rate of adversarial in the training process is also related to the robust performance. The PGD-AT  that adopts the adversarial examples with the highest attack success rate has the best robust performance. Compared with PGD2-AT, our FGSM-SDI  has a higher attack success rate and achieves a better robust performance. 
\par  {The generative network is one of the core parts of the proposed method. We adopt ResNet18 as the target model on CIFAR10 to explore the impact of the generative network. In detail, when training the generator, we perform m iterations on it, which can be dubbed FGSM-SDI-m. We record the robustness accuracy of adversarial examples
generated by PGD-10 on test data in the training phase. The robustness accuracy curves under the PGD-10 attack are shown in Fig.~\ref{fig:overfitting_iter}. It can be observed that improving the training iteration of the generator can improve the robustness performance, especially at the beginning of training. That indicates that model robustness increases as generator training progresses. And we also visualize the  adversarial initialization and FGSM-updated
perturbations for the FGSM-RS and our FGSM-SDI among continuous training
epochs. As shown in Fig.~\ref{fig:init_diff}, it can be observed that compared with the random initialization, the proposed initialization
is more informative. }
\par Adversarial perturbations fool a well-trained  model by interfering with important local regions that determine image classification. To explore whether our FGSM-SDI will be affected by adversarial perturbations, we adopt Gradient-weighted Class Activation Mapping (Grad-CAM)~\cite{DBLP:journals/ijcv/SelvarajuCDVPB20} to generate the heat maps that locate the category-related areas in the image. As shown in Fig.~\ref{fig:cam}, it can be observed that as for FGSM-RS,
adversarial perturbations modify the distribution of the maximal points on the generated heat map, while as for our FGSM-SDI, the adversarial perturbations do not  modify the distribution of the maximal points on the generated heat-map. That indicates that our FGSM-SDI is more robust. Moreover, we compare the loss landscape of the proposed method with those of the other fast AT methods to explore the association between latent hidden perturbation and local linearity. As shown in Fig~\ref{fig:loss_land}, compared with other AT methods, the cross-entropy loss of our FGSM-SDI is more linear in the adversarial direction. Using the  latent perturbation generated by the proposed method can preserve the local linearity of the target model better. 
It qualitatively proves that using the proposed   sample-dependent   adversarial   initialization can boost the fast AT.  {And to explore the stability of the proposed method FGSM-SDI, we train the proposed method multiple times and record the robustness accuracy of adversarial examples generated by PGD-10 on test data in the training phase. The mean and variance of robustness accuracy is shown in Fig~\ref{fig:mean_std}. It can be observed that the proposed method keeps stable robustness accuracy against adversarial examples. }

\section{Conclusion} \label{sec:conclusion}
In this paper, we propose a sample-dependent adversarial initialization to boost fast AT. Specifically, we adopt a generative network conditioned on a benign image and its gradient information from the target network to generate an effective initialization. 
In the training phase, the generative network and the target network are optimized jointly and play a game. 
The former learns to produce a dynamic sample-dependent initialization to generate stronger adversarial examples based on the current target network. 
And the latter adopts the generated adversarial examples for training to improve model robustness. 
Compared with widely adopted random initialization fashions in fast AT, the proposed initialization overcomes the catastrophic overfitting, thus improves model robustness.
Extensive experimental results demonstrate the superiority of our proposed method.




%
\IEEEpeerreviewmaketitle

\ifCLASSOPTIONcaptionsoff
  \newpage
\fi



\bibliographystyle{IEEEtran}
%

\bibliography{egbib_v3}

%








\begin{IEEEbiography}[{\includegraphics[width=0.8in,height=1in,clip,keepaspectratio]{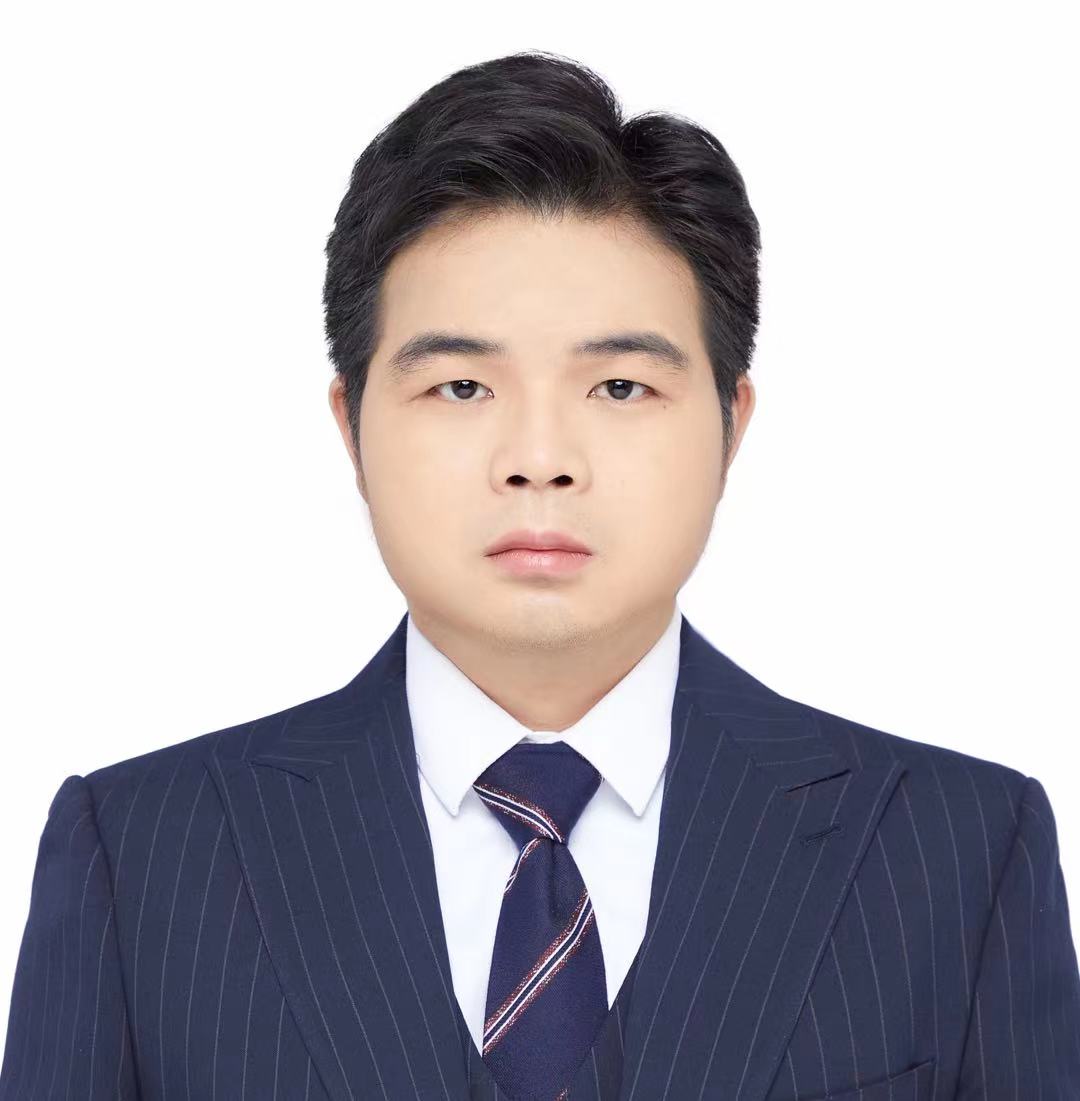}}]
{Xiaojun Jia} 
received his B.S. degree in Software Engineering from China University of Geosciences, China. He is now a Ph.D student in State Key Laboratory of Information Security, Institute
of Information Engineering, Chinese Academy of Sciences and School of
Cyber Security, University of Chinese Academy of Sciences, Beijing. His research interests include computer vision, deep learning and adversarial machine learning. He is the author of referred journals and conferences in IEEE CVPR, AAAI, ACM Multimedia etc.
\end{IEEEbiography}
\vspace{-6mm}
\begin{IEEEbiography}
[{\includegraphics[width=0.8in,height=1in,clip,keepaspectratio]{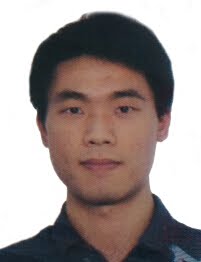}}]
{Yong Zhang} 
received the Ph.D. degree  in pattern recognition and intelligent systems from the Institute of Automation, Chinese Academy of Sciences in 2018. From 2015 to 2017, he was a Visiting Scholar with the Rensselaer Polytechnic Institute. He is currently with the Tencent AI Lab. His research interests include computer vision and machine learning.
\end{IEEEbiography}


\begin{IEEEbiography}
[{\includegraphics[width=0.8in,height=1in,clip,keepaspectratio]{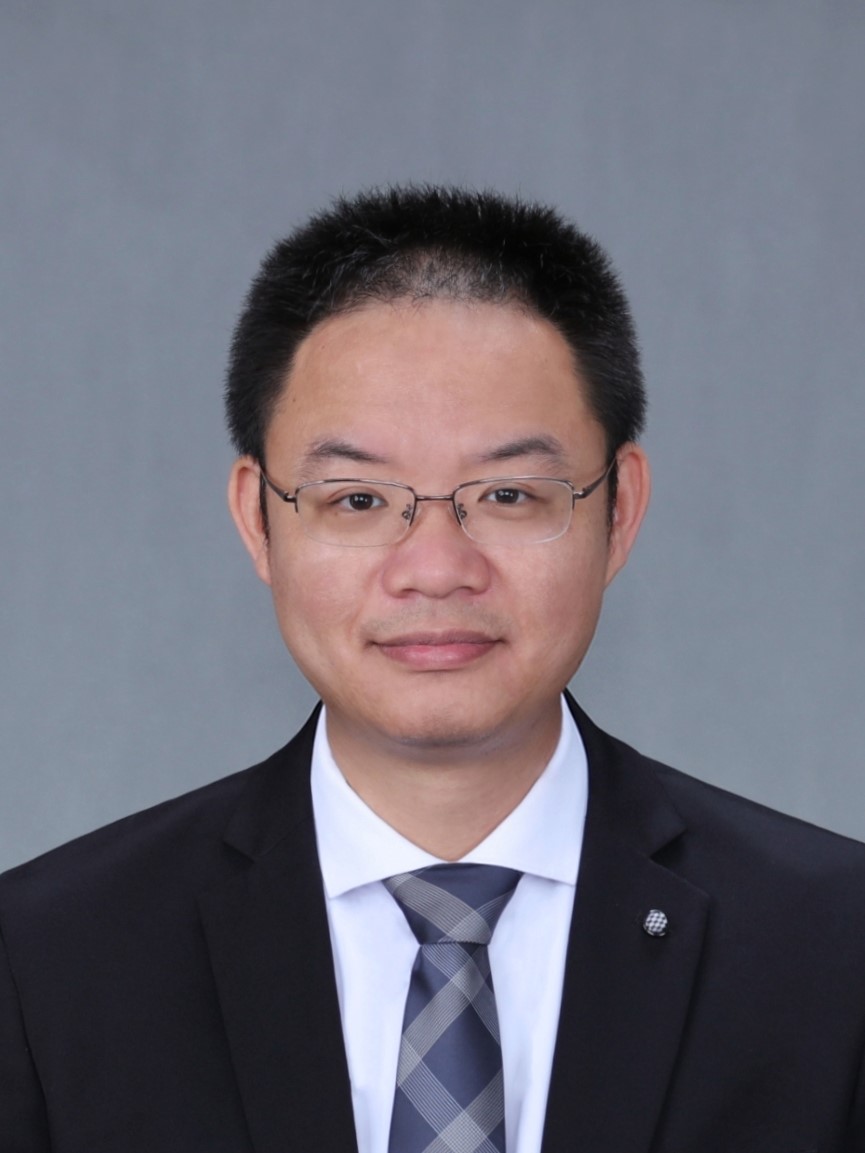}}]
{Baoyuan Wu} is an Associate Professor of School of Data Science, the Chinese University of Hong Kong, Shenzhen (CUHK-Shenzhen). He is also the director of the Secure Computing Lab of Big Data, Shenzhen Research Institute of Big Data (SBRID). On June 2014, he received the PhD degree from the National Laboratory of Pattern Recognition, Institute of Automation, Chinese Academy of Sciences. From November 2016 to August 2020, he was a Senior and Principal Researcher at Tencent AI lab. His research interests are AI security and privacy, machine learning, computer vision and optimization. He has published 40+ top-tier conference and journal papers, including TPAMI, IJCV, NeurIPS, CVPR, ICCV, ECCV, ICLR, AAAI, and one paper was selected as the Best Paper Finalist of CVPR 2019. He serves as an Associate Editor of Neurocomputing, Area Chair of ICLR 2022, AAAI 2022 and ICIG 2021, Senior Program Committee Member of AAAI 2021 and IJCAI 2020/2021, Task Force Member of CCF and CAA. He is the principal investigator of General Program of National Natural Science Foundation of China, 2021 CCF-Tencent Rhino-Bird Young Faculty Open Research Fund, and 2021 Tencent Rhino-Bird Special Research Fund.
\end{IEEEbiography}

\begin{IEEEbiography}
[{\includegraphics[width=0.8in,height=1in,clip,keepaspectratio]{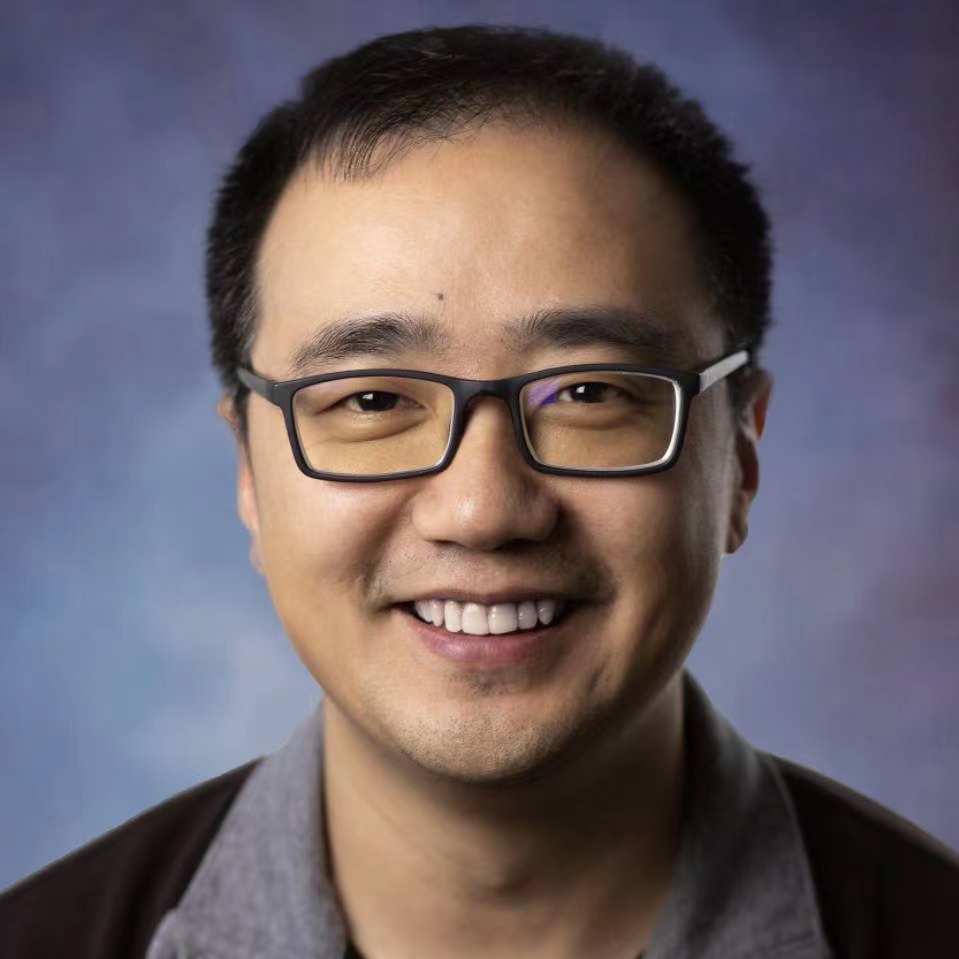}}]
{Jue Wang} received the B.E. and M.Sc. degrees from the Department of Automation, Tsinghua University, Beijing, China, and the Ph.D. degree in Electrical Engineering from the University of Washington in Seattle WA, USA. He is currently the Director of the Visual Computing Center at Tecent AI Lab. He was Senior Director at Megvii Research from 2017 to 2020. He was Principle Research Scientist at Adobe Research from 2007 to 2017. He has published more than 140 peer-reviewed research articles in the areas of Computer Vision, Computer Graphics and HCI,  and holds more than 60 international patents. He was a co-organizer for IEEE International Conference on Multimedia and Expo 2016 and IEEE International Conference on Computational Photography 2012.  He is Associated Editor for IEEE Transactions on Pattern Analysis and Machine Intelligence (T-PAMI), The Visual Computer and International Journal of Computer Games Technology. He was the recipient of the Microsoft Research Fellowship and the Yang Research Award of the University of Washington in 2006. He is a senior member of IEEE and ACM.
\end{IEEEbiography}

\begin{IEEEbiography}
[{\includegraphics[width=0.8in,height=1in,clip,keepaspectratio]{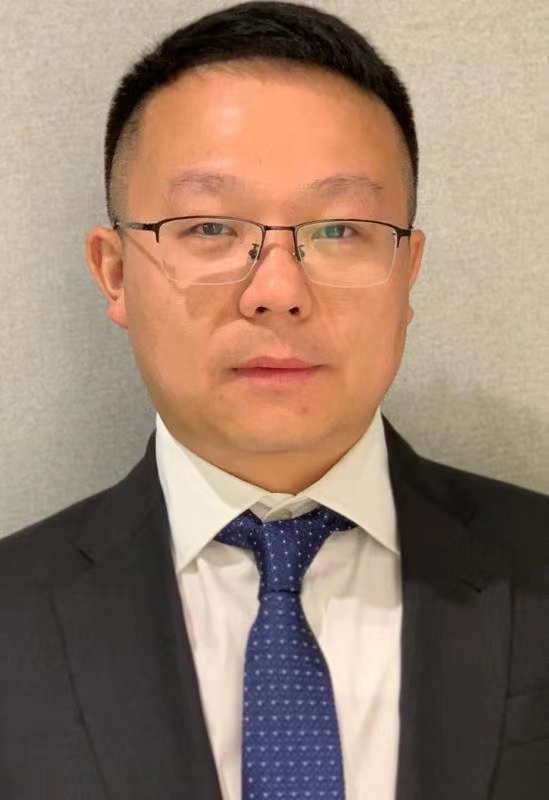}}]
{Xiaochun Cao}(SM'14)
received the B.S. and M.S. degrees in computer science from Beihang University, Beijing, China, and the Ph.D. degree in computer science from the University of Central Florida, Orlando, FL, USA. After graduation, he spent about three years at ObjectVideo Inc. as a Research Scientist. He is with School of Cyber Science and Technology, Shenzhen Campus, Sun Yat-sen University, Shenzhen 518107, P.R. China. He has authored and coauthored more than 100 journal and conference papers.
Prof. Cao is a Fellow of the IET. He is on the Editorial Boards of the IEEE Transactions on Image Processing, IEEE Transactions on Multimedia, IEEE Transactions on Circuits and Systems for Video Technology. His dissertation was nominated for the University of Central Florida's university-level Outstanding Dissertation Award. In 2004 and 2010, he was the recipient of the Piero Zamperoni Best Student Paper Award at the International Conference on Pattern Recognition.
\end{IEEEbiography}

\end{document}